\def\draft{1} % a flag for turning on/off authors' comments
\pdfoutput=1
\documentclass[11pt]{article}

\usepackage[preprint]{acl}

\usepackage{times}
\usepackage{latexsym}
\usepackage{amsfonts}
\usepackage{array}
\usepackage[T1]{fontenc}
\usepackage[utf8]{inputenc}

\usepackage{xcolor}
\usepackage{microtype}
\usepackage{tcolorbox}
\usepackage{amsmath}
\tcbuselibrary{skins}

\usepackage{inconsolata}

\usepackage{graphicx}
\usepackage{cleveref}

\title{The Geometry of Prompting: Unveiling Distinct Mechanisms of Task Adaptation in Language Models}

\author{
Artem Kirsanov \\
New York University \\
\texttt{kirsaa01@nyu.edu}
\\\And
Chi-Ning Chou \\
Flatiron Institute 
\\\And
Kyunghyun Cho \\
New York University \\
Genentech
\\\And
SueYeon Chung \\
New York University \\
Flatiron Institute
}

\newcommand{\Cnotes}[1]{\ifnum\draft=1{\color{red} [CNC: #1]}\fi}
\newcommand{\Anotes}[1]{\ifnum\draft=1{\color{blue} [AK: #1]}\fi}
\newcommand{\Snotes}[1]{\ifnum\draft=1{\color{magenta} [SYC: #1]}\fi}

\begin{document}
\maketitle

\begin{abstract}

Decoder-only language models have the ability to dynamically switch between various computational tasks based on input prompts. Despite many successful applications of prompting, there is very limited understanding of the internal mechanism behind such flexibility. In this work, we investigate how different prompting methods affect the geometry of representations in these models. Employing a framework grounded in statistical physics, we reveal that various prompting techniques, while achieving similar performance, operate through distinct representational mechanisms for task adaptation. Our analysis highlights the critical role of input distribution samples and label semantics in few-shot in-context learning. We also demonstrate evidence of synergistic and interfering interactions between different tasks on the representational level. Our work contributes to the theoretical understanding of large language models and lays the groundwork for developing more effective, representation-aware prompting strategies.

\end{abstract}

\section{Introduction}

A striking feature of modern language models (LMs) is their computational flexibility. Unlike traditional neural networks trained for specific tasks, LMs function as flexible computers that can be programmed (prompted) with natural language to perform a wide array of tasks. 

This adaptability, often termed in-context learning (ICL), has revolutionized natural language processing by enabling rapid task adaptation without expensive fine-tuning. However, despite ICL's widespread success, its underlying mechanisms remain poorly understood.

While some research has linked ICL to gradient-based learning \cite{vonoswald2023transformerslearnincontextgradient, akyürek2023learningalgorithmincontextlearning}, recent evidence in naturalistic settings suggests that ICL may not be pure "learning", but rather a method of steering the model to select familiar tasks from its pretraining corpus \cite{pan2023incontextlearninglearnsincontext, hendel2023incontextlearningcreatestask}. Recent studies have also highlighted importance of prompt design and demonstrated that the choice of examples and output labels can significantly impact performance \cite{zhao2021calibrateuseimprovingfewshot, min2022rethinkingroledemonstrationsmakes}. However, these works have primarily focused on the input-output behavior of LMs, leaving the internal dynamics of ICL largely unexplored.

In this work, we aim to illuminate ICL by investigating how different prompting methods modify internal representations in pre-trained language models. When a model is prompted to perform a classification task, we analyze the separability and geometric properties of category manifolds --- point clouds in the model's embedding space corresponding to examples sharing a category label. We leverage the recently developed framework of \textbf{manifold capacity} \cite{chung2018classification, chou2024neural}, which analytically connects task performance to the geometric properties of these representations.

Our core contributions are:

\begin{enumerate}

    \item A comprehensive analysis of how various prompting methods affect internal representations in language models, revealing distinct computational mechanisms despite similar performance outcomes.
    
    \item Novel insights into in-context learning dynamics, including the role of label semantics, synergistic effects of demonstrations on unrelated tasks, and representational trade-offs during task adaptation.
    
\end{enumerate}

\section{Related work}

\subsection{Prompting as task-adaptation}

The idea that a language model pretrained on next-token prediction can adapt to various tasks without parameter updates was popularized by \cite{brown2020languagemodelsfewshotlearners}. This phenomenon, known as in-context learning (ICL), relates to the model's ability to effectively "learn" a novel task by analogy from a few demonstration examples provided in the input sequence. To distinguish conventional few-shot ICL from other recently proposed input-based task-adaption methods, we refer to it as providing \textbf{demonstrations}, highlighting the crucial role of task examples.

While performance generally improves with more examples \cite{brown2020languagemodelsfewshotlearners, bertsch2024incontextlearninglongcontextmodels}, ICL exhibits counter-intuitive features, with performance being heavily dependent on the exact choice of examples, their ordering, formatting, and other factors \cite{zhao2021calibrateuseimprovingfewshot, wang2024largelanguagemodelslatent, liu2024understandingincontextlearningcontrastive}. Additionally, the actual input-output mapping matters less than expected \cite{min2022rethinkingroledemonstrationsmakes}, suggesting that few-shot ICL involves a complex interplay of true task learning from examples and task recognition from the pre-training corpus \cite{pan2023incontextlearninglearnsincontext}.

Language models also demonstrate zero-shot learning abilities, performing tasks based on abstract descriptions without explicit examples \cite{Radford2019LanguageMA, wei2022finetunedlanguagemodelszeroshot}. We refer to such task-adapting prompts without examples as \textbf{instructions}\footnote{We use "instruction" referring only to the format of the prompt for zero-shot learning and do all experiments on base models that were not instruction fine-tuned}. While often considered together under the umbrella of ICL, our results reveal that despite comparable performance, these two prompt types affect internal representations differently, highlighting the crucial role of input distribution examples.

Recently, prompt-tuning has emerged as an alternative approach to task adaptation \cite{lester2021powerscaleparameterefficientprompt, liu2022fewshotparameterefficientfinetuningbetter}. This method involves learning a small set of continuous vectors (soft prompts) that are concatenated to the input embeddings, while keeping the model parameters frozen. Prompt-tuning offers a middle ground between full model fine-tuning and static prompting, allowing for task-specific adaptations with significantly fewer trainable parameters.

\subsection{Internal representations}
Language computations rely on mapping individual words or tokens to vectors in a continuous embedding space, which possesses rich structure learned through model pretraining. The emerging \textit{linear representation hypothesis} \cite{park2024linearrepresentationhypothesisgeometry} suggests that this embedding space contains "feature directions" encoding human-interpretable concepts, allowing the model to perform vector operations with meaningful semantics \cite{mikolov2013efficientestimationwordrepresentations,pennington-etal-2014-glove, bowman2016generatingsentencescontinuousspace}.

The concept of feature superposition \cite{elhage2022toymodelssuperposition,arora2018linearalgebraicstructureword} provides insight into how a model can operate on more features than it has orthogonal directions in the embedding space. This is achieved by utilizing almost-orthogonal vectors for feature encoding with minimal interference, potentially circumvented by non-linear activation functions.

A popular method for uncovering encoded features involves training linear probes \cite{belinkov2021probingclassifierspromisesshortcomings} to "read out" information linearly from the embedding space. Probing methods have revealed the encoding of part-of-speech tags \cite{belinkov-etal-2017-neural}, parse-tree geometry \cite{hewitt-manning-2019-structural}, and higher-level semantic features such as spatial location of landmarks \cite{gurnee2024languagemodelsrepresentspace} and color \cite{abdou-etal-2021-language}. However, while these studies are usually performed on and averaged over a very diverse input corpus of text, there is a lack of understanding how the context preceding a given input (particularly, task adaptation) affects feature representation.

\subsection{Representational geometry}

The notion that underlying representations in the embedding space shape task performance has gained traction in both machine learning and computational neuroscience \cite{chungNeuralPopulationGeometry2021,fleschOrthogonalRepresentationsRobust2022, ansuiniIntrinsicDimensionData, fawziEmpiricalStudyTopology2018}. Intuitively, for a classification task, this implies that collective representations of inputs sharing a target category (a category manifold) must be well-separated from other categories. This concept of "manifold untangling" has been a prominent perspective on computational objectives in neuroscience \cite{dicarloUntanglingInvariantObject2007}.

The recently developed framework of manifold capacity \cite{chung2018classification, wakhloo2023linear, chou2024neural} proposes a formal link between representational geometry and separability. Manifold capacity quantifies how efficiently task-relevant features are encoded from the perspective of a linear downstream decoder. Essentially, it measures the separability of target classes in the embedding space, capturing the effectiveness of task-relevant feature encoding.

This framework has been successfully applied to investigate representational geometry in vision networks \cite{stephenson2019untangling, cohenSeparabilityGeometryObject2020, stephenson2021geometrygeneralizationmemorizationdeep} and language models \cite{mamou2020emergence}. By examining how different prompting methods affect manifold capacity, we can gain insights into the internal dynamics of ICL and the efficiency of various task adaptation strategies.

\section{Methods}

\subsection{Dataset details}

To investigate effects of various prompting methods on representations in different task-specific contexts, we required a dataset with control over multiple categorical dimensions of text. We could not find an existing text classification dataset with a sufficient number of samples and a comprehensive multilabel scheme suitable for tractable manifold analysis. Therefore, we leveraged a separate language model (Claude 3.5 Sonnet) to generate a synthetic dataset tailored to our research requirements. This synthetic dataset consists of diverse sentences, each simultaneously labeled with three types of categories: Sentiment, Topic, and Intent, with five categories for each type. Such multidimensional labeling allowed us to investigate representational effects in a multitasking setting (see sections \ref{multitasking} and \ref{prompt-tuning}).

For consistency, all experiments, including those focused on single-task performance (section \ref{ICL-single-task}), utilized this dataset, with the sentiment classification task serving as our primary focus. To validate our findings, we also replicated key single-task experiments using established open datasets as a control.

Full details on the datasets, including generation process, category distributions, and example sentences, are provided in the \cref{sec:appendix_dataset_details}.

\subsection{Task setup}

Our work focuses on text  classification tasks with a fixed set of categories, as such tasks have an analytically-grounded link between the geometry of underlying representation and separability of categories in the embedding space, ultimately determining the end performance. 

In contrast to traditional encoder-based models, where separate linear classifiers are trained to predict target category directly from the embedding vectors, we investigate decoder-only language models. These models can prompted to generate class labels directly in the vocabulary space.

This approach introduces two key factors affecting performance:
\begin{enumerate}
\item \textbf{Representation Quality}: The underlying representation in the embedding space must support the separation of class manifolds.
\item \textbf{Readout Alignment}: The alignment between the model's unembed layer and the ideal decoder directions impacts the final output quality.
\end{enumerate}

\begin{figure}[h!]
    \centering
    \includegraphics[width=1\linewidth]{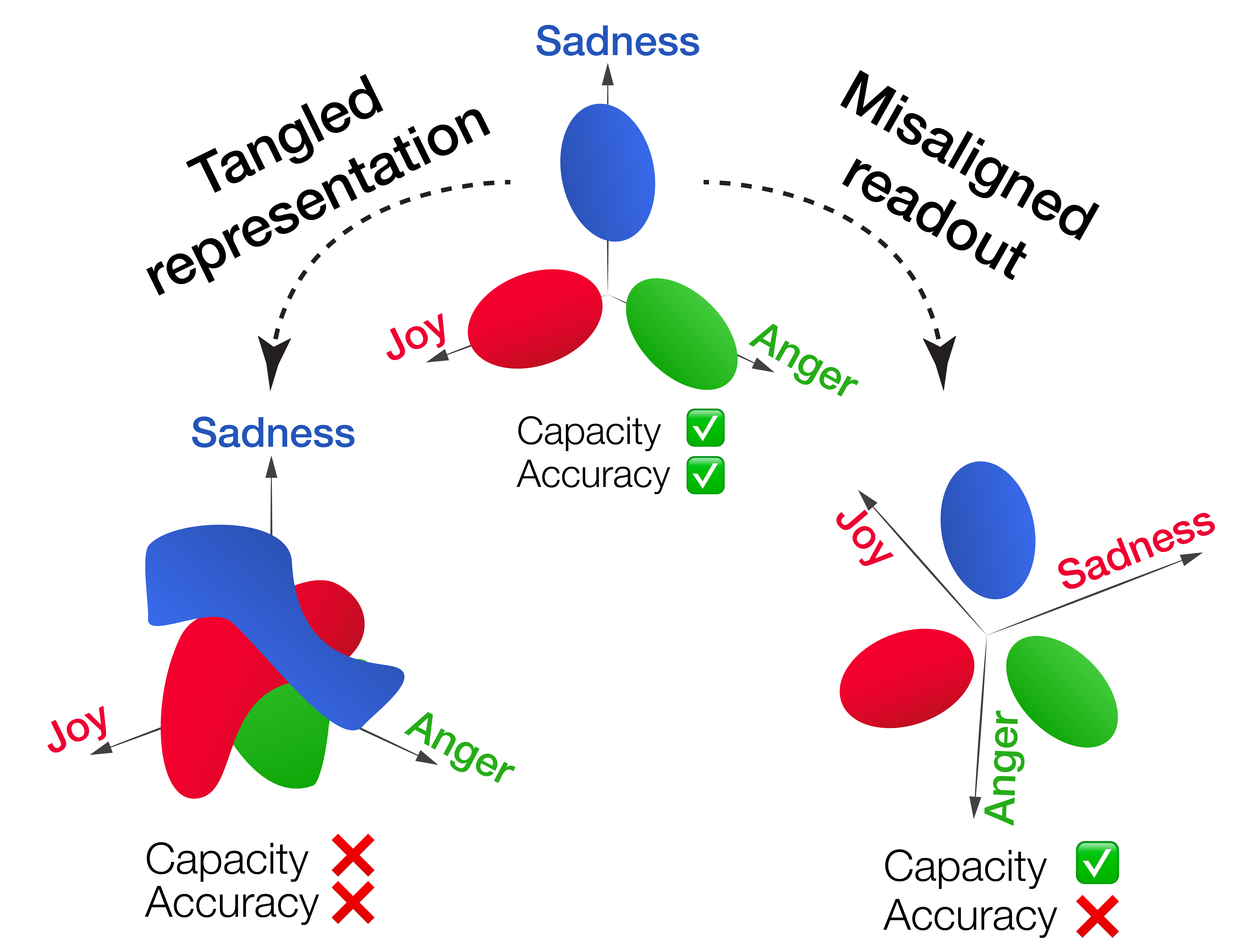}
    \caption{Two components of the model’s performance. Low accuracy can be caused by either suboptimal and tangled representation in the embedding space (left), as well as misalignment between the representation and model’s readout layer (right). Manifold capacity, which relates the performance of an ideal decoder to the underlying geometry can differentiate between the 2 cases.}
    
    % ~\Cnotes{Maybe add texts in the figure on something like manifold untangle (check or not), readout alignment (check or not)}
    
    \label{fig:readout_representation_schematics}
\end{figure}

Manifold capacity theory allows us to disentangle these components by quantifying the representation quality at each layer, independently of the specific unembed module being used for vocabulary readout. This idea is schematically illustrated in \cref{fig:readout_representation_schematics}.

\subsection{Prompting strategies}
Throughout the work we compared two main types of natural-language prompting. \textbf{Instruction} prompt consisted of the following text (using sentiment as an example): 

\begin{tcolorbox}[  enhanced,
  interior style={
    top color=gray!5,
    bottom color=gray!5,
  },
  frame style={
    color=gray!35,
  },
  left=2pt,    % left padding
  right=2pt,   % right padding
  top=2pt,     % top padding
  bottom=2pt,  % bottom padding
]
  This is a text classification task. Possible categories are Joy, Sadness, Fear, Anger, Surprise.
  
  Text: \textit{[Test Sentence]}
  
  Category:
\end{tcolorbox}

where \textit{[Test Sentence]} stands for the sentence text from the dataset that is being evaluated.

\textbf{Demonstration} prompt consisted of a variable number of examples following a similar format:

\begin{tcolorbox}[
  enhanced,
  interior style={
    top color=blue!2,
    bottom color=blue!5,
  },
  frame style={
    left color=blue!10,
    right color=blue!10,
    middle color=blue!10
  },
  left=2pt,    % left padding
  right=2pt,   % right padding
  top=2pt,     % top padding
  bottom=2pt,  % bottom padding
]
Text: \textit{[Demo sentence 1]}

Category: Joy

Text: \textit{[Demo sentence N]}

Category: Fear

Text: \textit{Test Sentence}

Category:
\end{tcolorbox}

As a baseline control for the representation analysis, we also extracted embedding using the \textbf{raw sentence} input of the following format:

\begin{tcolorbox}[  enhanced,
  interior style={
    top color=gray!2,
    bottom color=gray!2,
  },
  frame style={
    color=gray!10,
  },
  left=2pt,    % left padding
  right=2pt,   % right padding
  top=2pt,     % top padding
  bottom=2pt,  % bottom padding
]
  Text: \textit{[Test Sentence]}
  
  Category:
\end{tcolorbox}

\subsection{Embedding Extraction}
Analyzing representational geometry in decoder-only models presents unique challenges due to masked self-attention, distributed sentence-level features, and last token dependency. To address these challenges and investigate the effects of prompting on representations, we consider two types of embeddings:
\begin{enumerate}

\item \textbf{Sentence Embeddings}: We extract and residual stream activations for tokens corresponding only to the input sentence, excluding the task prompt, and average their embedding vectors along sequence dimension. This provides insight into the model's intermediate processing stage.

\item\textbf{Last-token Embeddings}: We extract residual stream activations of the last token in the sequence at each layer. This allows us to track how sentence-level features are aggregated into the final representation used for output generation.

\end{enumerate}

\begin{figure}[h!]
    \centering
    \includegraphics[width=1\linewidth]{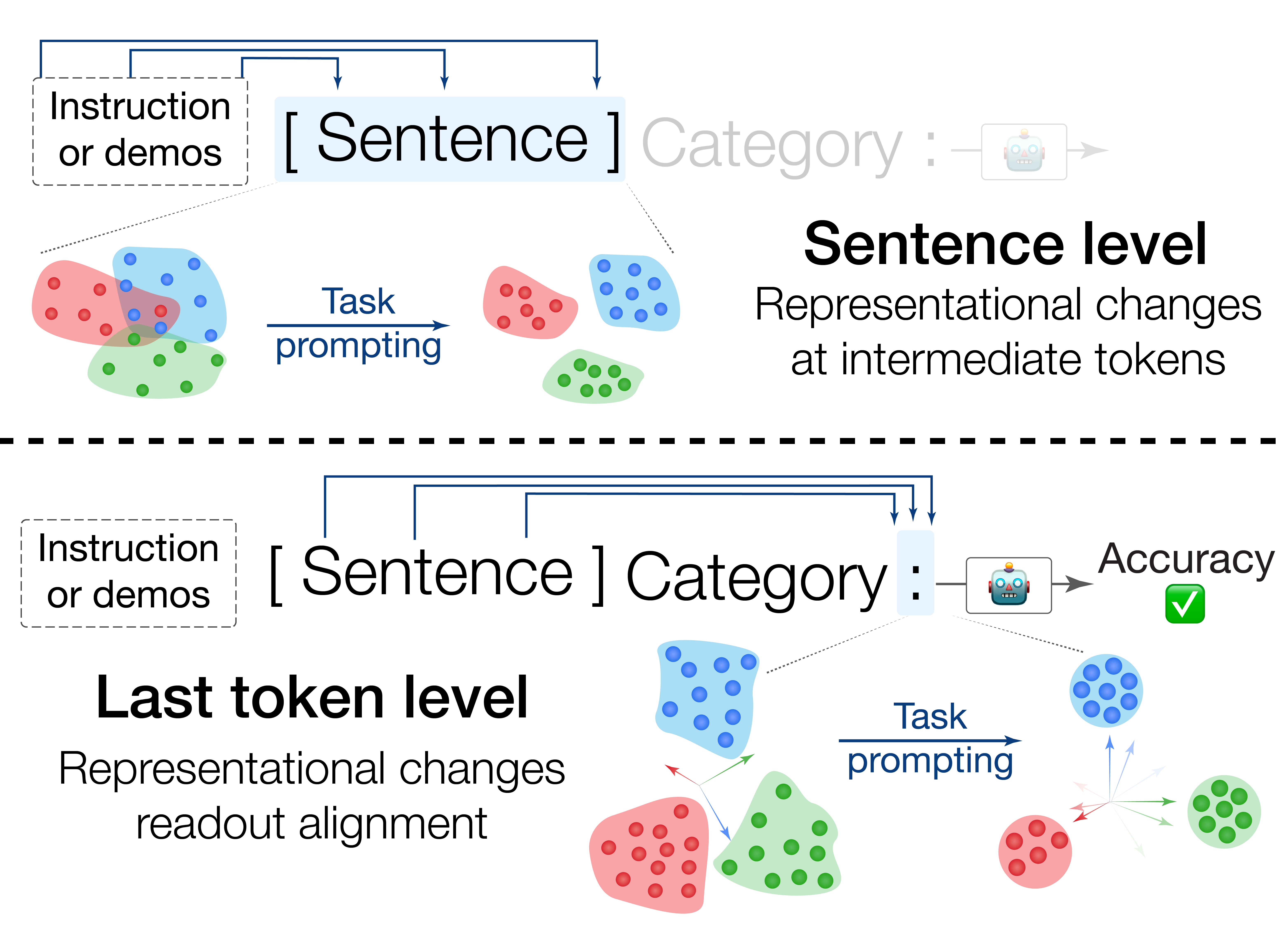}
    \caption{Possible effect sites of prompting. Task-specific prefix might affect extraction of relevant features at the sentence-level, reorganizing intermediate representations (top). High performance would also imply more efficient repackaging of extracted features into the embedding of the last token, as well readout alignment (bottom).}
    \label{fig:two_embeddings_schematics}
\end{figure}

These embedding types and possible effects of prompting are illustrated in \cref{fig:two_embeddings_schematics}.

\subsection{Analysis of representations}

To analyse representational geometry we first construct category manifolds (point clouds) by accumulating the embedding vectors of all sentences sharing a class label. So for a classification task with $P$ categories, the resulting representation can be thought of as $P$ distinct collections of vectors in the embedding space. We then compute the following properties of the resulting collective representation.

\paragraph{Manifold capacity} 

Capacity is a positive scalar measure, that measures how separable the underlying category manifolds are, with higher capacity values corresponding to higher degree of separability. Intuitively, it can be thought of as the "number of linearly decodable classes per dimension", quantifying how efficiently manifolds are packed in the embedding space \cite{gardner1988optimal, chung2018classification}. We provide a more mathematically detailed explanation of one interpretation of capacity in \cref{appendix:manifold_capacity}, and for full rigorous treatment, refer the reader to  \cite{chou2024neural}.

\paragraph{Geometry of individual manifolds}

In this work we make a few simplifications, compared to the original formulation \cite{chou2024neural}.  Manifold capacity is analytically expressed as a function of so called \textit{effective} radius and dimension of manifolds, that are determined by the spatial arrangement of manifolds' anchor points, that can be thought of as support vectors for the classification problem. In the presence of correlated structure, these measures might have complicated form, not necessary corresponding to intuitive notions of radius and dimension. To bring our results into a more direct interpretation, we measure geometry in the following way instead:

\begin{enumerate}
    \item \textbf{Dimension} of each manifold was measured as participation ratio of principal components,  which roughly corresponds the number of dimensions needed to explain around 80--90\% of total variance \cite{gaoTheoryMultineuronalDimensionality2017}.
    \item \textbf{Radius} of each manifold was taken to be the maximum distance between any pair of points on the manifold.
\end{enumerate}

Both metrics were averaged across manifolds, each resulting in a single scalar value. We refer to these measures as geometric properties of individual manifolds, since they do not depend on the relative positions and orientations of manifolds in the embedding space.

\paragraph{Correlation structure}

Manifold capacity also depends on the spatial arrangement of individual manifolds relative to each other and to the global origin. We measure correlation coefficients between axes of variation of individual manifolds (\textbf{Axes-alignment}) and correlations between each manifold's axes and its centroid (\textbf{Center-axes alignment}). For an extended discussion of how these correlation measures affect manifold capacity in different regimes, see \cite{chou2024neural}. In this work, we consider these measures collectively as \textit{correlation structure} to explain capacity changes driven by the relative arrangements of manifolds in the embedding space, rather than by changes in individual manifold properties.

\section{Results}

Our analysis reveals complex dynamics in how prompting affects the internal representations of language models, with distinct patterns emerging at different processing stages and for various prompting methods.

\subsection{Representational changes during text-classification task}
\label{ICL-single-task}

We first investigated performance and representational effects of prompting during a conventional ICL setting, comparing demonstrations and instruction prompts.

\begin{figure}[h!]
    \centering
    \includegraphics[width=1\linewidth]{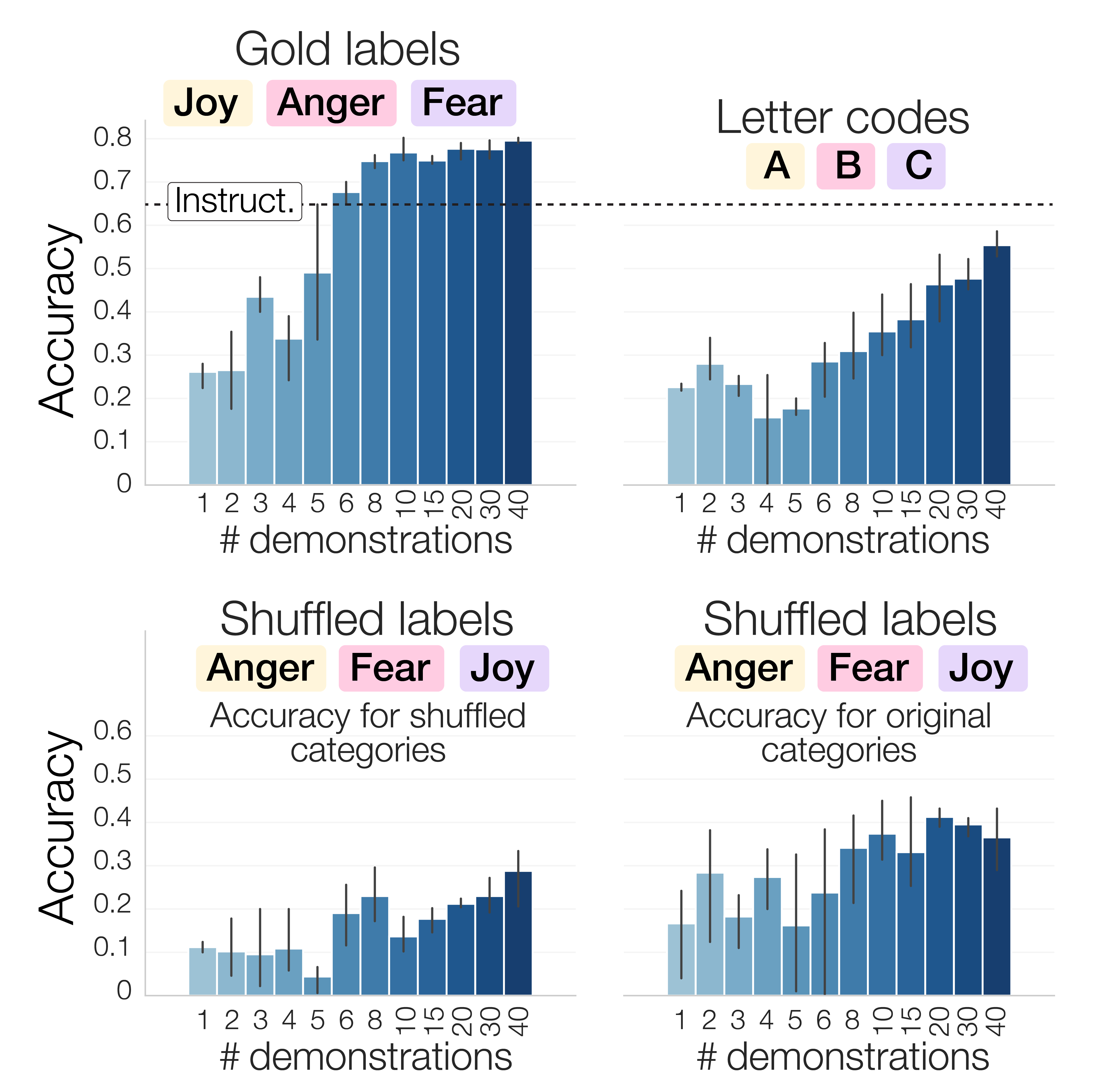}
    \caption{Performance of demonstrations and instruction prompting on sentiment analysis task.}
    \label{fig:performance_barplots_emotion}
\end{figure}

\paragraph{Task performance} 
Instruction alone achieved good accuracy, outperforming demonstration prompts with few examples ($\leq 5$). Larger example sets ($>5$) surpassed explicit instruction, with performance quickly plateauing (\cref{fig:performance_barplots_emotion}). Replacing meaningful category words (gold labels) with abstract letters required more demonstration examples to infer category nature.  When category labels were consistently shuffled (e.g. "Joy" $\rightarrow$ "Anger"), the model failed to generalize beyond pretrained associations, achieving low accuracy for both target (shuffled) and original labels. This suggests that the model is not purely learning a novel task from scratch, but rather (at least partially) relies on existing associations encoded in label semantics. \cite{pan2023incontextlearninglearnsincontext}.

\paragraph{Sentence-level effects}

\begin{figure}[h!]
    \centering
    \includegraphics[width=1\linewidth]{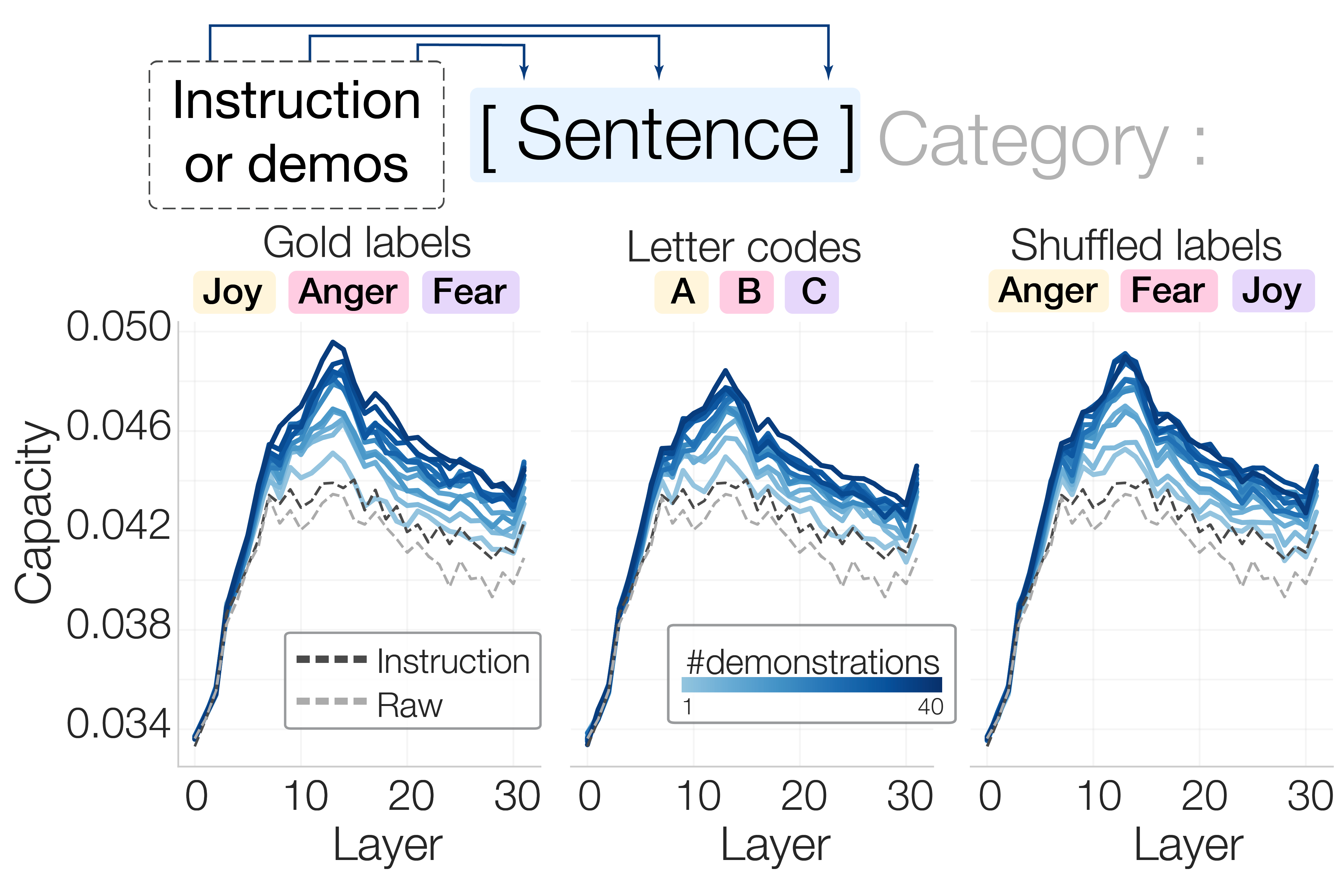}
    \caption{Manifold capacity of \textbf{sentence-level} embeddings during demonstrations prompting compared to instruction and raw sentence control}
    \label{fig:geometry_mean_pooled_emotion}
\end{figure}

Analysis of sentence-level embeddings (\cref{fig:geometry_mean_pooled_emotion}) revealed that demonstration examples, but not abstract instruction, significantly reorganized intermediate representations at early-mid layers. This reorganization increased the separability of sentiment manifolds by reducing manifold dimension and improving correlation structure (see \cref{fig:ICL_sentence_level_extended}). Surprisingly, there was little difference in resulting geometry between demonstrations across three labeling strategies, indicating that sentence representation is primarily influenced by input distribution examples, rather than input-output mapping.

\paragraph{Last-token effects}

\begin{figure}[h!]
    \centering
    \includegraphics[width=1\linewidth]{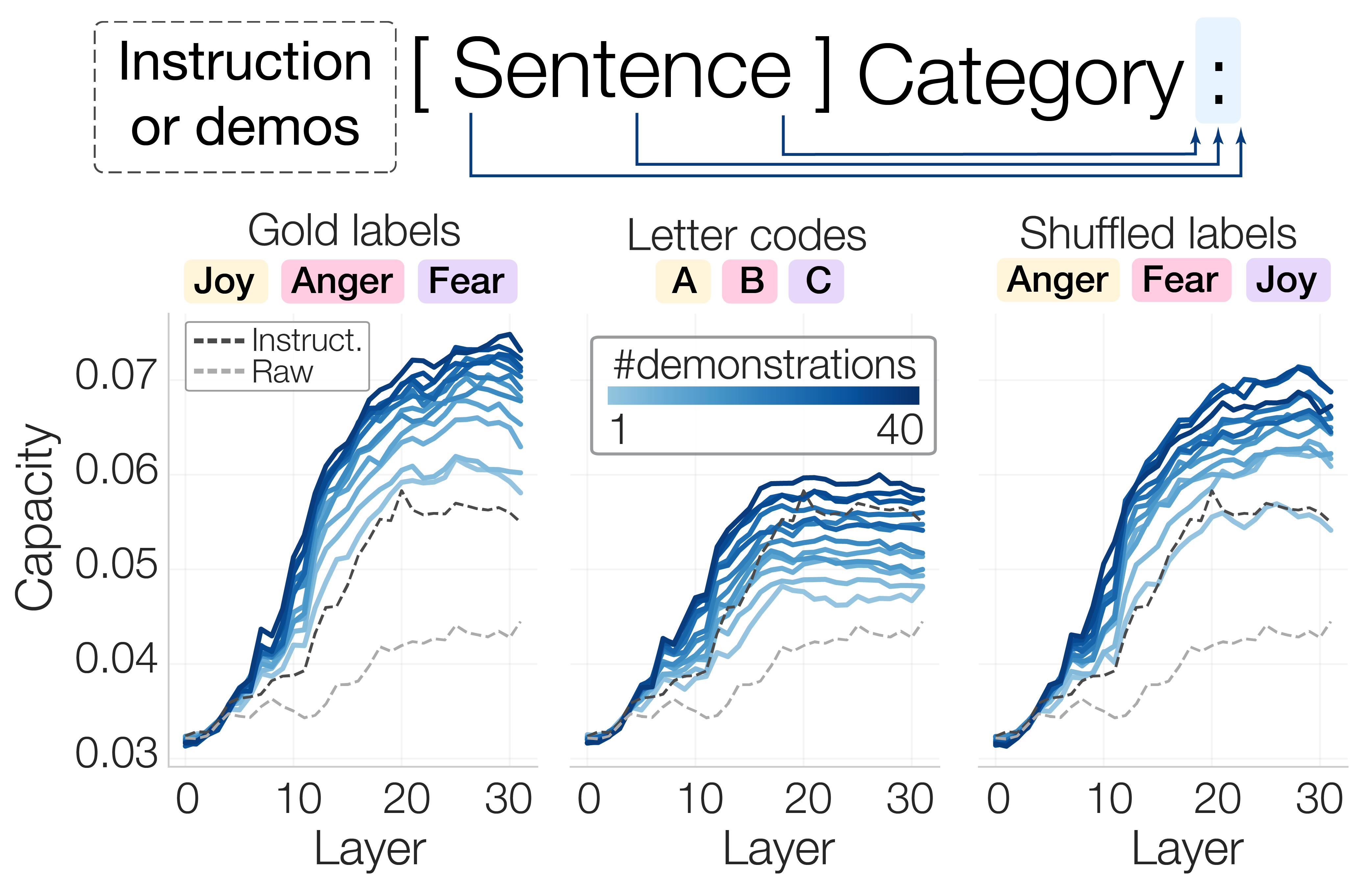}
    \caption{Manifold capacity of \textbf{last token} embeddings during demonstrations prompting compared to instruction and raw sentence control.}
    \label{fig:geometry_last_token_emotion}
\end{figure}

At the last-token level, instruction prompts significantly increased manifold capacity relative to raw sentences, with effects emerging as early as layer 8 and persisting to final layers (\cref{fig:geometry_last_token_emotion}). Geometrically, the increased separability was driven mostly by the reduction in dimension along with correlation structure (supplementary \cref{fig:ICL_last_token_extended}). Demonstrations further increased manifold capacity compared to instruction, despite lower task performance for cases with few examples. This suggests that while instruction alone achieves better accuracy due to high alignment between the model's readout and category manifolds, demonstrations improve both readout alignment and representation structure. Even just for a couple of demonstrations, the underlying representation is already more optimal compared to instruction-prompted case, but this separability is not utilized properly by the unembed layer. Notably, last-token capacity during letter code labeling was much lower compared to category words, explaining lower performance when output labels lack meaningful semantics. For shuffled labels capacity values were similar to the gold label setting, suggesting that model's inability to overwrite existing associations is explained by the readout misalignment, while the underlying representation is intact.
 
\paragraph{Sensitivity to the choice of demonstrations}

\begin{figure}[h!]
    \centering
    \includegraphics[width=1\linewidth]{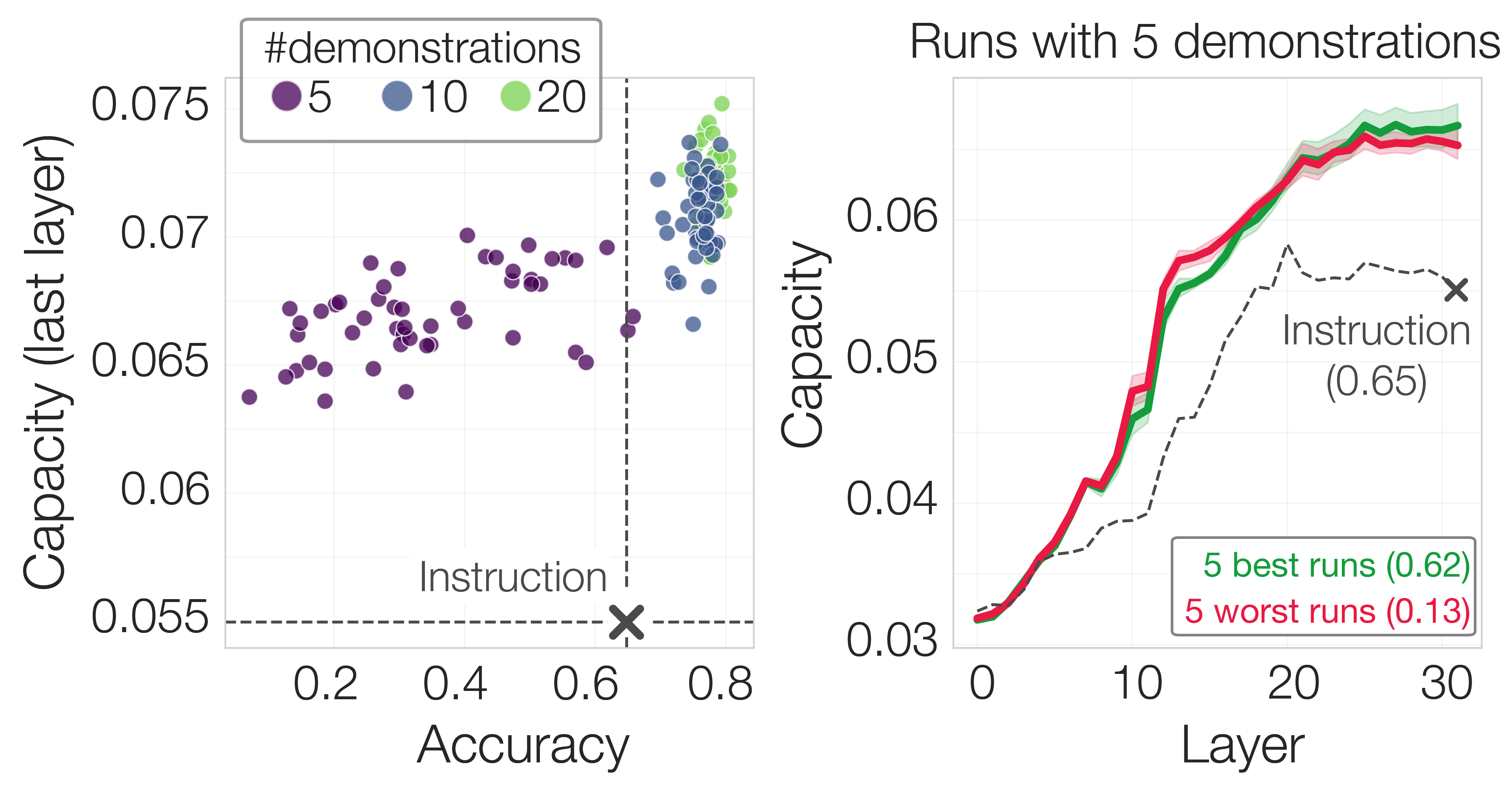}
    \caption{Left: Manifold capacity at the final layer versus accuracy for individual ICL runs with different numbers of demonstrations compared to instruction. Right: Layer-wise capacity profiles for the five best- and worst-performing 5-demonstration runs compared to instruction (accuracies shown in brackets)}
    \label{fig:ICL_stability}
\end{figure}

Performance of few-shot ICL has been previously reported to depend heavily on the choice of particular examples and their ordering, even for a fixed number of demonstrations provided \cite{zhao2021calibrateuseimprovingfewshot}. To investigate whether such failure modes of certain training sets stem from changes in the underlying geometry, we analyzed the relationship between last-token manifold capacity and end performance across multiple random samplings of demonstrations, while keeping the number of examples fixed. In accordance with prior work, we observed large variance in performance (\cref{fig:ICL_stability} left), particularly in settings with fewer examples. For instance, with five demonstrations, accuracy varied dramatically from below 0.1 to approximately 0.6. Despite this substantial performance variability, the changes in manifold capacity of the last token embedding at the final layer were minimal. Even in "failure" runs with lowest accuracy, manifold capacity was significantly higher than in the instruction setting, and the layer-wise profile of capacity in the worst runs was nearly identical to the best runs (\cref{fig:ICL_stability} right). These results provide further evidence that the instability of few-shot ICL and its sensitivity to particular examples is driven primarily by poor readout alignment rather than differences in representational geometry

\subsection{Cross-Task Interactions in Multi-Task Prompting}

\label{multitasking}

\paragraph{Multi-task setup} We then investigated whether prompting a model to perform one task would affect the quality of representation for another unrelated task. To this end, we constructed an artificial sentence-classification dataset containing three independent sets of labels, each with five categories. This design allowed each sentence to be classified by its sentiment, topic, or intent (see \cref{sec:appendix_dataset_details} for details).

We created instruction and demonstration prompts for each of the three tasks. We then computed representational metrics for each of the three possible sets of manifolds, resulting in nine possible pairs between a prompt and a representation. We termed the three cases where the manifold-inducing labels coincided with the classification objective (e.g., performing sentiment analysis and computing separability of sentiment manifolds) as \textbf{\textcolor[HTML]{2f7139}{coherent}}. The remaining six cases, where manifold capacity was evaluated for a different set of labels, were termed \textbf{\textcolor[HTML]{9e281e}{incoherent}}.

This setup allowed us to explore how prompting for one task affects the model's internal representations not just for that task, but also for other potential tasks on the same input. All experiments used gold category labels, and manifold metrics for each configuration were normalized by the corresponding value in the raw sentence case.

\paragraph{Synergistic Effects at the Sentence Level} 

\begin{figure}[h!]
    \centering
    \includegraphics[width=1\linewidth]{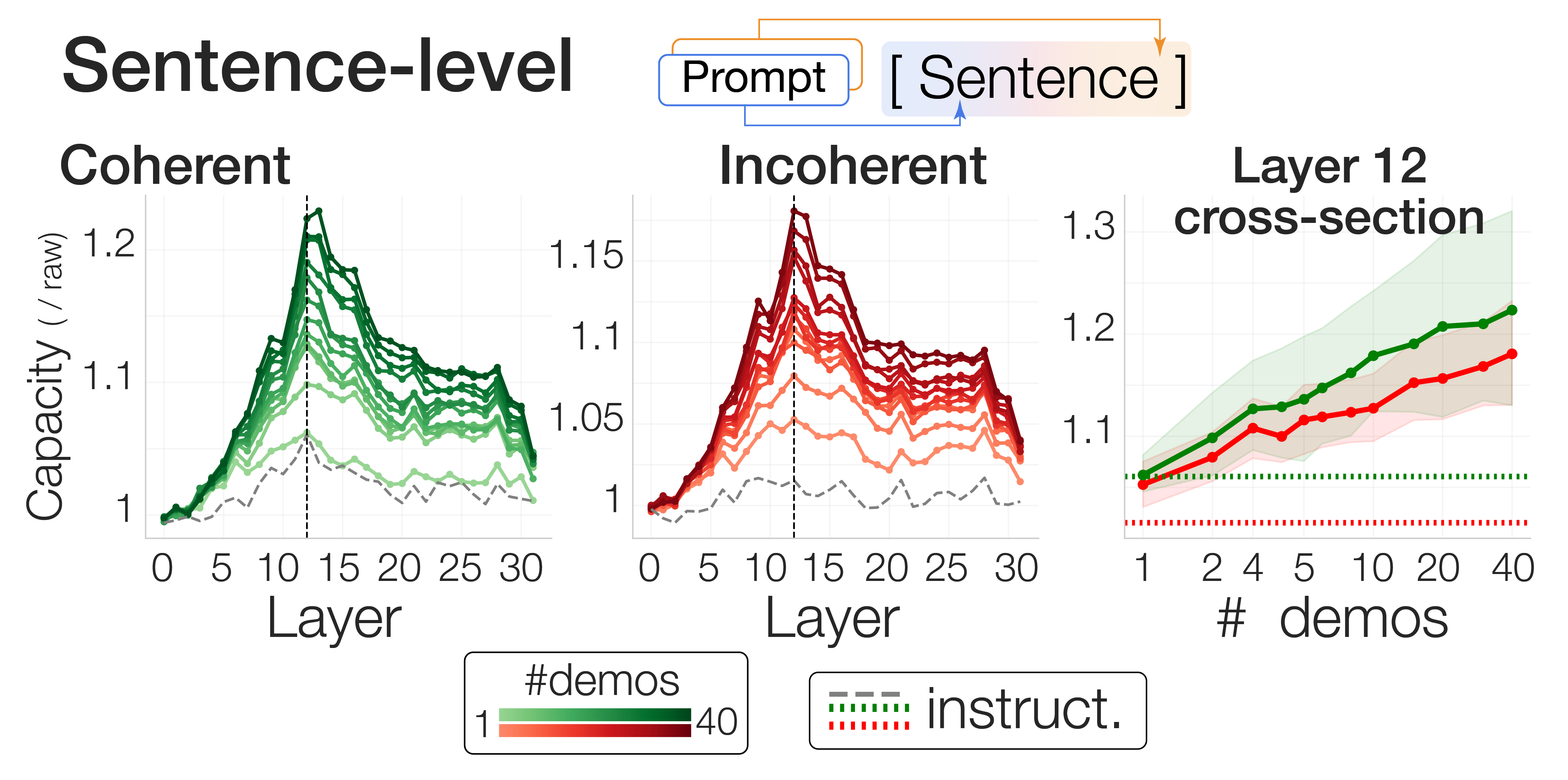}
    \caption{Effect of prompting in a multitask setting at sentence-level}
    \label{fig:multitasking_sentence_level}
\end{figure}

Increasing the number of demonstrations robustly led to increased manifold capacity at intermediate layers for coherent configurations, while instruction had a much weaker effect (\cref{fig:multitasking_sentence_level}). Surprisingly, demonstrations for an incoherent task also increased capacity with a similar layerwise profile, albeit to a lesser extent. This highlights the role of input distribution: providing example sentences enhances representation capacity for supporting other tasks on the same input distribution, even in the context of a different task. While the trend of increased capacity with growing number of examples was similar for both coherent and incoherent scenarios, the amplitude of such increase was larger when the task was coherent with the manifold labels. Notably, while the overall trend is captured by the decrease in manifold dimension, the difference between coherent and incoherent settings is not fully explained by the geometry of individual manifolds (supplementary \cref{fig:multitask_extended_sentence}). Instead, it likely arises due to changes in the correlation structure and relative positions of manifolds in the embedding space.

\paragraph{Task Interference in Last Token Representations} 
 
\begin{figure}[h!]
    \centering
    \includegraphics[width=1\linewidth]{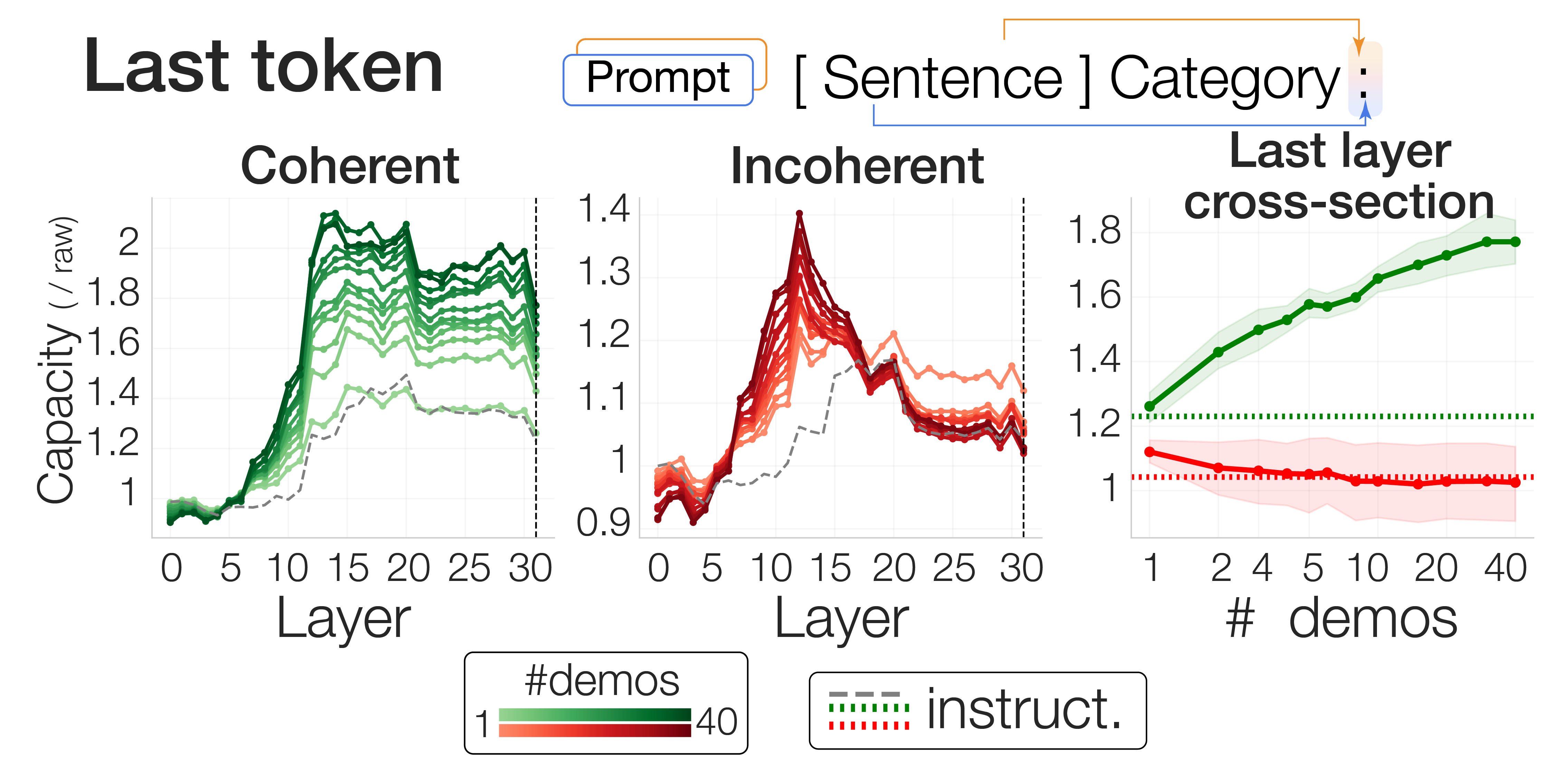}
    \caption{Effect of prompting in a multitask setting at last-token level}
    \label{fig:multitasking_last_token}
\end{figure}

Analysis of last token embeddings revealed an interesting dichotomy of layerwise dynamics (\cref{fig:multitasking_last_token}). At earlier layers, additional demonstrations of incoherent tasks increased manifold capacity, but at later layers, this trend reversed, with additional examples decreasing capacity. Coherent demonstrations significantly increased capacity starting with layer 12 and persisting to the final layer. The increase in capacity driven by coherent prompts at intermediate layers was much more prominent, compared to incoherent prompts, indicating a larger role of task-specific input-output pairings. Decrease in capacity at final layers with growing number of demonstrations suggests an intriguing idea of representational tradeoff:  as the model prepares the output, features for irrelevant tasks, that were emphasized at intermediate processing stages are compromised in favor of better separability of task-relevant features. Interestingly, this effect could not be explained by the geometry of individual manifolds --- we observed a reduction in dimension with increased number of examples for both coherent and incoherent tasks. Instead, we observed that center-axes correlations behaved differently for coherent and incoherent cases, capturing the trend in capacity (see supplementary \cref{fig:multitask_extended_last_token}).

\subsection{Distinct representational mechanisms of prompt-tuning}

\label{prompt-tuning}
\begin{figure*}
    \centering
    \includegraphics[width=1\linewidth]{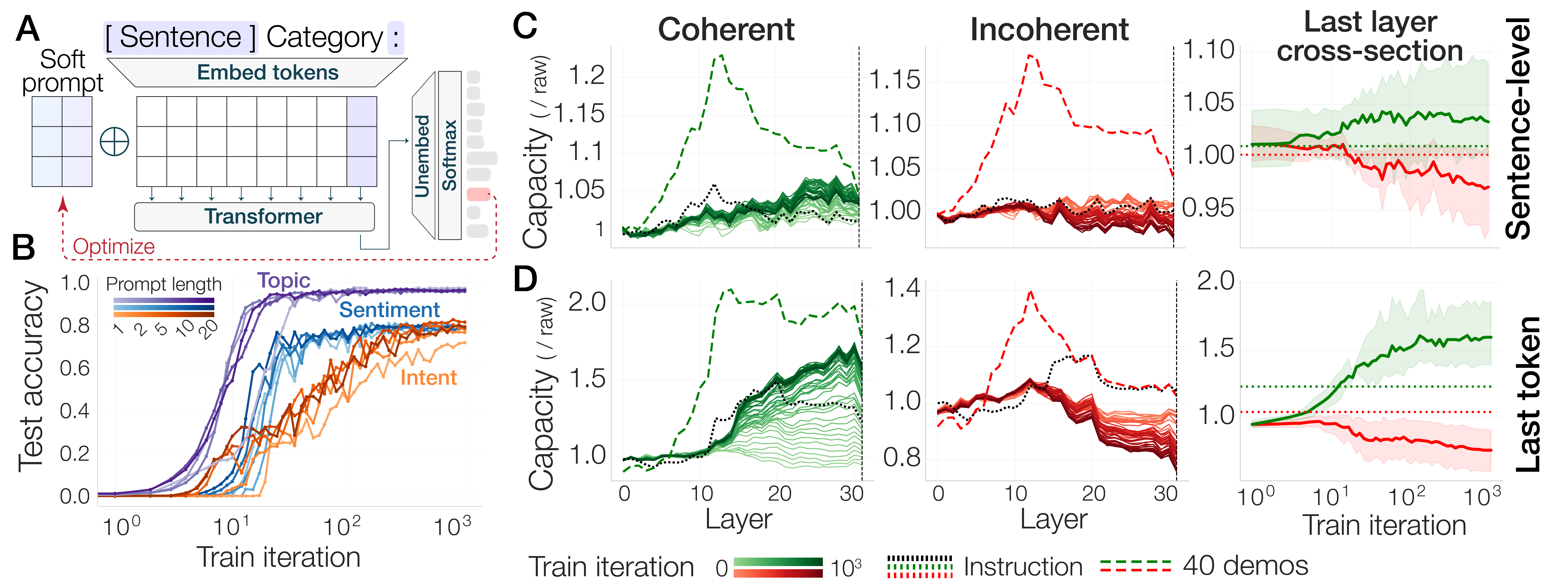}
    \caption{Schematic of the prompt-tuning setup (A) and performance at various tasks for different lengths of the soft prompt (B). Right: Manifold capacity changes during training across layers for sentence-level (C) and last token (D) representations}
    \label{fig:prompt_tuning}
\end{figure*}

\paragraph{Performance and Setup of Soft Prompts}

Finally, we extended our investigation to \textbf{prompt-tuning}, an alternative method of task adaptation \cite{lester2021powerscaleparameterefficientprompt} that optimizes a task-specific prompt directly in the embedding space (hence "soft"), which is prepended to the test input (\cref{fig:prompt_tuning} A). This approach allowed us to examine whether gradient-based methods for task adaptation affect internal representations similarly to traditional prompting methods.

To validate performance, we trained separate soft prompts of varying lengths for all three tasks, computing test accuracy across intermediate training iterations (\cref{fig:prompt_tuning}, B). Notably, soft prompts consistently outperformed demonstrations ((see \ref{sec:appendix}), and we found no significant correlation between final performance and prompt length, consistent with prior work \cite{lester2021powerscaleparameterefficientprompt}. 
Given the similarity in performance and representational effects across different prompt lengths, we present representative plots for a 5-token soft prompt in the following analysis.

\paragraph{Minimal Impact on Intermediate Representations}  Analysis of sentence-level representations during soft prompt training revealed a striking mechanistic difference compared to hard natural-language prompts (\cref{fig:prompt_tuning} C). The optimization-based solution did not alter intermediate representations in earlier layers, as illustrated by the absence of the characteristic peak around layer 12 observed with other prompting methods. Instead, effects were concentrated in later layers, where we observed a similar representational trade-off even at the sentence level: prompts for incoherent tasks led to decreased capacity. Importantly, this capacity difference could not be explained by the geometry of individual manifolds, suggesting the critical role of relative manifold arrangement  (see \cref{fig:prompt_tuning_extended_sentence}).

\paragraph{Enhanced Trade-off in Last Token Representations} 

At the last token level, prompt-tuning also exhibited distinct effects. For coherent prompts, manifold capacity increased substantially in later layers as training progressed, surpassing instruction-prompted capacity but remaining below that induced by demonstrations. Notably, this effect emerged at later layers compared to both instruction and demonstration methods. In the incoherent case, soft prompts dramatically reduced the capacity of representations for unrelated tasks. This suggests that gradient-based input optimization compromises the representation of task-irrelevant features even more than natural-language demonstrations. As with demonstrations, the capacity difference between coherent and incoherent settings was primarily attributable to the relative alignment of manifolds in the embedding space, rather than geometry of individual manifolds ( \cref{fig:prompt_tuning_extended_last_tooken}).

Taken together, our results on prompt-tuning indicate that soft-prompts, often proposed to be alternative to demonstration-based ICL, operate through fundamentally different internal mechanisms compared to demonstrations and zero-shot instruction.

\section{Discussion}

Our study illuminates the mechanisms of how language models adapt to various tasks by analyzing the geometry of internal representations under different prompting methods. We found that zero-shot instruction, few-shot demonstrations, and tunable soft-prompts, while achieving comparable performance, operate through distinctly different representational mechanisms.

Zero-shot instructions, while effective, primarily influence the final stages of processing, affecting how features are "packaged" in the last token embedding without significantly altering intermediate representations. In contrast, demonstration examples have a more profound impact, reshaping intermediate representations to optimize them for the classification objective. In a multitask setting, demonstrations optimize early-layer representations to support multiple potential tasks, regardless of the specific task being demonstrated. This suggests a form of general feature enhancement triggered by exposure to diverse input examples. Soft-prompts, despite being trained on the same input distribution examples, operate differently, mainly affecting later layers responsible for output preparation, distinguishing them from the broader impact of natural language demonstrations.

A key insight emerging from our analysis is the distinction between representational geometry and readout alignment in determining model performance. Manifold capacity measures the inherent separability of category representations and their potential for supporting robust classification across all possible linear readouts. However, actual model performance also depends on a specific readout -- the model's unembed layer --- which may fail to optimally utilize well-structured representations. This effect manifests itself in two notable "failure modes" of few-shot ICL --- dramatic sensitivity to the choice and ordering of specific examples and the inability to generalize beyond label associations in the pretraining corpus. In both cases, the internal representations remain well-organized for classification, but the unembed layer fails to effectively leverage this structure, resulting in poor accuracy. High separability, as measured by manifold capacity, suggests that one could train a simple linear readout module on top of existing representations to overcome this, leveraging the feature-extraction power of decoder-only LLMs for efficiently adapting their representations to specific tasks. The success of prompt-tuning further supports this view: its effectiveness appears to stem primarily from improving the alignment between representations and the vocabulary readout layer, rather than fundamentally altering the geometric organization of the embedding space.

These findings suggest two promising directions for future research. First, given that internal representations often maintain high manifold capacity even when ICL performance is poor, there is significant potential in better understanding and optimizing readout alignment. Quantifying decoder alignment by comparing the performance of independently trained classifiers with the model's unembed layer could provide deeper insights into this bottleneck and suggest ways to overcome it. Second, our observation that demonstrations can drastically change representational geometry suggests opportunities for more direct geometric optimization. Recent work has shown promising results in related fields: optimizing vision network parameters to directly maximize manifold separability has achieved SoTA performance \cite{yerxa2023learning}, while regularizing learned embeddings to respect structural characteristics has improved performance in causal inference tasks \cite{balashankar-subramanian-2021-learning}. We anticipate that similar insights into LM representational geometry could drive innovations in language model prompting, enhancing performance and stability across a wide range of objectives.

\section*{Limitations}

Our study provides insights into the representational geometry of language models under different prompting methods, but it has limitations. First, we used synthetic datasets generated by Claude 3.5 Sonnet, which allowed precise control over task parameters. However, this approach may not fully capture the complexity and variability of real-world language structure. To enhance the generalizability of our findings, future research should expand testing to include a broader range of natural datasets.

Second, the metrics used to quantify representational geometry in our study, such as manifold capacity and individual manifold geometry, though informative, simplify the more complex tasks that occur in language models, by focusing on a classification task with given target labels. Future work should examine how other tasks, such as those requiring multi-token outputs (e.g., chain-of-thought prompting), affect representational geometry. Additionally, more advanced measures that link geometry to complex computations could provide further insights into the fine-grained changes during task adaptation.

\section*{Acknowledgments}

This work was funded by the Center for Computational Neuroscience at the Flatiron Institute of the Simons Foundation. S.C. is supported by the Klingenstein-Simons Award, a Sloan Research Fellowship, NIH award R01DA059220, and the Samsung Advanced Institute of Technology (under the project "Next Generation Deep Learning: From Pattern Recognition to AI"). All experiments were performed on the Flatiron Institute's high-performance computing cluster.

\bibliography{acl_latex}

\begin{thebibliography}{47}
\providecommand{\natexlab}[1]{#1}

\bibitem[{Abdou et~al.(2021)Abdou, Kulmizev, Hershcovich, Frank, Pavlick, and S{\o}gaard}]{abdou-etal-2021-language}
Mostafa Abdou, Artur Kulmizev, Daniel Hershcovich, Stella Frank, Ellie Pavlick, and Anders S{\o}gaard. 2021.
\newblock \href {https://doi.org/10.18653/v1/2021.conll-1.9} {Can language models encode perceptual structure without grounding? a case study in color}.
\newblock pages 109--132, Online. Association for Computational Linguistics.

\bibitem[{Akyürek et~al.(2023)Akyürek, Schuurmans, Andreas, Ma, and Zhou}]{akyürek2023learningalgorithmincontextlearning}
Ekin Akyürek, Dale Schuurmans, Jacob Andreas, Tengyu Ma, and Denny Zhou. 2023.
\newblock \href {https://arxiv.org/abs/2211.15661} {What learning algorithm is in-context learning? investigations with linear models}.
\newblock \emph{Preprint}, arXiv:2211.15661.

\bibitem[{Ansuini et~al.(2019)Ansuini, Laio, Macke, and Zoccolan}]{ansuiniIntrinsicDimensionData}
Alessio Ansuini, Alessandro Laio, Jakob~H Macke, and Davide Zoccolan. 2019.
\newblock Intrinsic dimension of data representations in deep neural networks.

\bibitem[{Arora et~al.(2018)Arora, Li, Liang, Ma, and Risteski}]{arora2018linearalgebraicstructureword}
Sanjeev Arora, Yuanzhi Li, Yingyu Liang, Tengyu Ma, and Andrej Risteski. 2018.
\newblock \href {https://arxiv.org/abs/1601.03764} {Linear algebraic structure of word senses, with applications to polysemy}.
\newblock \emph{Preprint}, arXiv:1601.03764.

\bibitem[{Balashankar and Subramanian(2021)}]{balashankar-subramanian-2021-learning}
Ananth Balashankar and Lakshminarayanan Subramanian. 2021.
\newblock \href {https://doi.org/10.18653/v1/2021.acl-long.69} {Learning faithful representations of causal graphs}.
\newblock pages 839--850, Online. Association for Computational Linguistics.

\bibitem[{Belinkov(2021)}]{belinkov2021probingclassifierspromisesshortcomings}
Yonatan Belinkov. 2021.
\newblock \href {https://arxiv.org/abs/2102.12452} {Probing classifiers: Promises, shortcomings, and advances}.
\newblock \emph{Preprint}, arXiv:2102.12452.

\bibitem[{Belinkov et~al.(2017)Belinkov, Durrani, Dalvi, Sajjad, and Glass}]{belinkov-etal-2017-neural}
Yonatan Belinkov, Nadir Durrani, Fahim Dalvi, Hassan Sajjad, and James Glass. 2017.
\newblock \href {https://doi.org/10.18653/v1/P17-1080} {What do neural machine translation models learn about morphology?}
\newblock pages 861--872, Vancouver, Canada. Association for Computational Linguistics.

\bibitem[{Bertsch et~al.(2024)Bertsch, Ivgi, Alon, Berant, Gormley, and Neubig}]{bertsch2024incontextlearninglongcontextmodels}
Amanda Bertsch, Maor Ivgi, Uri Alon, Jonathan Berant, Matthew~R. Gormley, and Graham Neubig. 2024.
\newblock \href {https://arxiv.org/abs/2405.00200} {In-context learning with long-context models: An in-depth exploration}.
\newblock \emph{Preprint}, arXiv:2405.00200.

\bibitem[{Bowman et~al.(2016)Bowman, Vilnis, Vinyals, Dai, Jozefowicz, and Bengio}]{bowman2016generatingsentencescontinuousspace}
Samuel~R. Bowman, Luke Vilnis, Oriol Vinyals, Andrew~M. Dai, Rafal Jozefowicz, and Samy Bengio. 2016.
\newblock \href {https://arxiv.org/abs/1511.06349} {Generating sentences from a continuous space}.
\newblock \emph{Preprint}, arXiv:1511.06349.

\bibitem[{Brown et~al.(2020)Brown, Mann, Ryder, Subbiah, Kaplan, Dhariwal, Neelakantan, Shyam, Sastry, Askell, Agarwal, Herbert-Voss, Krueger, Henighan, Child, Ramesh, Ziegler, Wu, Winter, Hesse, Chen, Sigler, Litwin, Gray, Chess, Clark, Berner, McCandlish, Radford, Sutskever, and Amodei}]{brown2020languagemodelsfewshotlearners}
Tom~B. Brown, Benjamin Mann, Nick Ryder, Melanie Subbiah, Jared Kaplan, Prafulla Dhariwal, Arvind Neelakantan, Pranav Shyam, Girish Sastry, Amanda Askell, Sandhini Agarwal, Ariel Herbert-Voss, Gretchen Krueger, Tom Henighan, Rewon Child, Aditya Ramesh, Daniel~M. Ziegler, Jeffrey Wu, Clemens Winter, Christopher Hesse, Mark Chen, Eric Sigler, Mateusz Litwin, Scott Gray, Benjamin Chess, Jack Clark, Christopher Berner, Sam McCandlish, Alec Radford, Ilya Sutskever, and Dario Amodei. 2020.
\newblock \href {https://arxiv.org/abs/2005.14165} {Language models are few-shot learners}.
\newblock \emph{Preprint}, arXiv:2005.14165.

\bibitem[{Chou et~al.(2024)Chou, Arend, Wakhloo, Kim, Slatton, and Chung}]{chou2024neural}
Chi-Ning Chou, Luke Arend, Albert~J Wakhloo, Royoung Kim, Will Slatton, and SueYeon Chung. 2024.
\newblock Neural manifold capacity captures representation geometry, correlations, and task-efficiency across species and behaviors.
\newblock \emph{bioRxiv}.

\bibitem[{Chung and Abbott(2021)}]{chungNeuralPopulationGeometry2021}
SueYeon Chung and L.F. Abbott. 2021.
\newblock \href {https://doi.org/10.1016/j.conb.2021.10.010} {Neural population geometry: {{An}} approach for understanding biological and artificial neural networks}.
\newblock \emph{Current Opinion in Neurobiology}, 70:137--144.

\bibitem[{Chung et~al.(2018)Chung, Lee, and Sompolinsky}]{chung2018classification}
SueYeon Chung, Daniel~D Lee, and Haim Sompolinsky. 2018.
\newblock Classification and geometry of general perceptual manifolds.
\newblock \emph{Physical Review X}.

\bibitem[{Cohen et~al.(2020)Cohen, Chung, Lee, and Sompolinsky}]{cohenSeparabilityGeometryObject2020}
Uri Cohen, SueYeon Chung, Daniel~D. Lee, and Haim Sompolinsky. 2020.
\newblock \href {https://doi.org/10.1038/s41467-020-14578-5} {Separability and geometry of object manifolds in deep neural networks}.
\newblock \emph{Nature Communications}, 11(1):746.

\bibitem[{DiCarlo and Cox(2007)}]{dicarloUntanglingInvariantObject2007}
James~J. DiCarlo and David~D. Cox. 2007.
\newblock \href {https://doi.org/10.1016/j.tics.2007.06.010} {Untangling invariant object recognition}.
\newblock \emph{Trends in Cognitive Sciences}, 11(8):333--341.

\bibitem[{Elhage et~al.(2022)Elhage, Hume, Olsson, Schiefer, Henighan, Kravec, Hatfield-Dodds, Lasenby, Drain, Chen, Grosse, McCandlish, Kaplan, Amodei, Wattenberg, and Olah}]{elhage2022toymodelssuperposition}
Nelson Elhage, Tristan Hume, Catherine Olsson, Nicholas Schiefer, Tom Henighan, Shauna Kravec, Zac Hatfield-Dodds, Robert Lasenby, Dawn Drain, Carol Chen, Roger Grosse, Sam McCandlish, Jared Kaplan, Dario Amodei, Martin Wattenberg, and Christopher Olah. 2022.
\newblock \href {https://arxiv.org/abs/2209.10652} {Toy models of superposition}.
\newblock \emph{Preprint}, arXiv:2209.10652.

\bibitem[{Fawzi et~al.(2018)Fawzi, {Moosavi-Dezfooli}, Frossard, and Soatto}]{fawziEmpiricalStudyTopology2018}
Alhussein Fawzi, Seyed-Mohsen {Moosavi-Dezfooli}, Pascal Frossard, and Stefano Soatto. 2018.
\newblock \href {https://doi.org/10.1109/CVPR.2018.00396} {Empirical {{Study}} of the {{Topology}} and {{Geometry}} of {{Deep Networks}}}.
\newblock In \emph{2018 {{IEEE}}/{{CVF Conference}} on {{Computer Vision}} and {{Pattern Recognition}}}, pages 3762--3770, Salt Lake City, UT. IEEE.

\bibitem[{Flesch et~al.(2022)Flesch, Juechems, Dumbalska, Saxe, and Summerfield}]{fleschOrthogonalRepresentationsRobust2022}
Timo Flesch, Keno Juechems, Tsvetomira Dumbalska, Andrew Saxe, and Christopher Summerfield. 2022.
\newblock \href {https://doi.org/10.1016/j.neuron.2022.01.005} {Orthogonal representations for robust context-dependent task performance in brains and neural networks}.
\newblock \emph{Neuron}, 110(7):1258--1270.e11.

\bibitem[{Gao et~al.(2017)Gao, Trautmann, Yu, Santhanam, Ryu, Shenoy, and Ganguli}]{gaoTheoryMultineuronalDimensionality2017}
Peiran Gao, Eric Trautmann, Byron Yu, Gopal Santhanam, Stephen Ryu, Krishna Shenoy, and Surya Ganguli. 2017.
\newblock \href {https://doi.org/10.1101/214262} {A theory of multineuronal dimensionality, dynamics and measurement}.

\bibitem[{Gardner and Derrida(1988)}]{gardner1988optimal}
Elizabeth Gardner and Bernard Derrida. 1988.
\newblock Optimal storage properties of neural network models.
\newblock \emph{Journal of Physics A: Mathematical and general}, 21(1):271.

\bibitem[{{Gemma Team}(2024)}]{gemmateam2024gemma2improvingopen}
{Gemma Team}. 2024.
\newblock \href {https://arxiv.org/abs/2408.00118} {Gemma 2: Improving open language models at a practical size}.
\newblock \emph{Preprint}, arXiv:2408.00118.

\bibitem[{Gurnee and Tegmark(2024)}]{gurnee2024languagemodelsrepresentspace}
Wes Gurnee and Max Tegmark. 2024.
\newblock \href {https://arxiv.org/abs/2310.02207} {Language models represent space and time}.
\newblock \emph{Preprint}, arXiv:2310.02207.

\bibitem[{Hendel et~al.(2023)Hendel, Geva, and Globerson}]{hendel2023incontextlearningcreatestask}
Roee Hendel, Mor Geva, and Amir Globerson. 2023.
\newblock \href {https://arxiv.org/abs/2310.15916} {In-context learning creates task vectors}.
\newblock \emph{Preprint}, arXiv:2310.15916.

\bibitem[{Hewitt and Manning(2019)}]{hewitt-manning-2019-structural}
John Hewitt and Christopher~D. Manning. 2019.
\newblock \href {https://doi.org/10.18653/v1/N19-1419} {{A} structural probe for finding syntax in word representations}.
\newblock pages 4129--4138, Minneapolis, Minnesota. Association for Computational Linguistics.

\bibitem[{Hovy et~al.(2001)Hovy, Gerber, Hermjakob, Lin, and Ravichandran}]{hovy-etal-2001-toward}
Eduard Hovy, Laurie Gerber, Ulf Hermjakob, Chin-Yew Lin, and Deepak Ravichandran. 2001.
\newblock \href {https://aclanthology.org/H01-1069} {Toward semantics-based answer pinpointing}.

\bibitem[{Kingma and Ba(2017)}]{kingma2017adammethodstochasticoptimization}
Diederik~P. Kingma and Jimmy Ba. 2017.
\newblock \href {https://arxiv.org/abs/1412.6980} {Adam: A method for stochastic optimization}.
\newblock \emph{Preprint}, arXiv:1412.6980.

\bibitem[{Lester et~al.(2021)Lester, Al-Rfou, and Constant}]{lester2021powerscaleparameterefficientprompt}
Brian Lester, Rami Al-Rfou, and Noah Constant. 2021.
\newblock \href {https://arxiv.org/abs/2104.08691} {The power of scale for parameter-efficient prompt tuning}.
\newblock \emph{Preprint}, arXiv:2104.08691.

\bibitem[{Li and Roth(2002)}]{li-roth-2002-learning}
Xin Li and Dan Roth. 2002.
\newblock \href {https://aclanthology.org/C02-1150} {Learning question classifiers}.

\bibitem[{Liu et~al.(2024)Liu, Xu, Li, Feng, and Song}]{liu2024understandingincontextlearningcontrastive}
Fuxiao Liu, Paiheng Xu, Zongxia Li, Yue Feng, and Hyemi Song. 2024.
\newblock \href {https://arxiv.org/abs/2307.05052} {Towards understanding in-context learning with contrastive demonstrations and saliency maps}.
\newblock \emph{Preprint}, arXiv:2307.05052.

\bibitem[{Liu et~al.(2022)Liu, Tam, Muqeeth, Mohta, Huang, Bansal, and Raffel}]{liu2022fewshotparameterefficientfinetuningbetter}
Haokun Liu, Derek Tam, Mohammed Muqeeth, Jay Mohta, Tenghao Huang, Mohit Bansal, and Colin Raffel. 2022.
\newblock \href {https://arxiv.org/abs/2205.05638} {Few-shot parameter-efficient fine-tuning is better and cheaper than in-context learning}.
\newblock \emph{Preprint}, arXiv:2205.05638.

\bibitem[{{Llama Team, AI @ Meta}(2024)}]{dubey2024llama3herdmodels}
{Llama Team, AI @ Meta}. 2024.
\newblock \href {https://arxiv.org/abs/2407.21783} {The llama 3 herd of models}.
\newblock \emph{Preprint}, arXiv:2407.21783.

\bibitem[{Mamou et~al.(2020)Mamou, Le, Del~Rio, Stephenson, Tang, Kim, and Chung}]{mamou2020emergence}
Jonathan Mamou, Hang Le, Miguel Del~Rio, Cory Stephenson, Hanlin Tang, Yoon Kim, and SueYeon Chung. 2020.
\newblock Emergence of separable manifolds in deep language representations.
\newblock In \emph{International Conference on Machine Learning}, pages 6713--6723. PMLR.

\bibitem[{Mikolov et~al.(2013)Mikolov, Chen, Corrado, and Dean}]{mikolov2013efficientestimationwordrepresentations}
Tomas Mikolov, Kai Chen, Greg Corrado, and Jeffrey Dean. 2013.
\newblock \href {https://arxiv.org/abs/1301.3781} {Efficient estimation of word representations in vector space}.
\newblock \emph{Preprint}, arXiv:1301.3781.

\bibitem[{Min et~al.(2022)Min, Lyu, Holtzman, Artetxe, Lewis, Hajishirzi, and Zettlemoyer}]{min2022rethinkingroledemonstrationsmakes}
Sewon Min, Xinxi Lyu, Ari Holtzman, Mikel Artetxe, Mike Lewis, Hannaneh Hajishirzi, and Luke Zettlemoyer. 2022.
\newblock \href {https://arxiv.org/abs/2202.12837} {Rethinking the role of demonstrations: What makes in-context learning work?}
\newblock \emph{Preprint}, arXiv:2202.12837.

\bibitem[{Pan et~al.(2023)Pan, Gao, Chen, and Chen}]{pan2023incontextlearninglearnsincontext}
Jane Pan, Tianyu Gao, Howard Chen, and Danqi Chen. 2023.
\newblock \href {https://arxiv.org/abs/2305.09731} {What in-context learning "learns" in-context: Disentangling task recognition and task learning}.
\newblock \emph{Preprint}, arXiv:2305.09731.

\bibitem[{Park et~al.(2024)Park, Choe, and Veitch}]{park2024linearrepresentationhypothesisgeometry}
Kiho Park, Yo~Joong Choe, and Victor Veitch. 2024.
\newblock \href {https://arxiv.org/abs/2311.03658} {The linear representation hypothesis and the geometry of large language models}.
\newblock \emph{Preprint}, arXiv:2311.03658.

\bibitem[{Pennington et~al.(2014)Pennington, Socher, and Manning}]{pennington-etal-2014-glove}
Jeffrey Pennington, Richard Socher, and Christopher Manning. 2014.
\newblock \href {https://doi.org/10.3115/v1/D14-1162} {{G}lo{V}e: Global vectors for word representation}.
\newblock pages 1532--1543, Doha, Qatar. Association for Computational Linguistics.

\bibitem[{Radford et~al.(2019)Radford, Wu, Child, Luan, Amodei, and Sutskever}]{Radford2019LanguageMA}
Alec Radford, Jeff Wu, Rewon Child, David Luan, Dario Amodei, and Ilya Sutskever. 2019.
\newblock \href {https://api.semanticscholar.org/CorpusID:160025533} {Language models are unsupervised multitask learners}.

\bibitem[{Stephenson et~al.(2019)Stephenson, Feather, Padhy, Elibol, Tang, McDermott, and Chung}]{stephenson2019untangling}
Cory Stephenson, Jenelle Feather, Suchismita Padhy, Oguz Elibol, Hanlin Tang, Josh McDermott, and SueYeon Chung. 2019.
\newblock Untangling in invariant speech recognition.
\newblock \emph{Advances in neural information processing systems}, 32.

\bibitem[{Stephenson et~al.(2021)Stephenson, Padhy, Ganesh, Hui, Tang, and Chung}]{stephenson2021geometrygeneralizationmemorizationdeep}
Cory Stephenson, Suchismita Padhy, Abhinav Ganesh, Yue Hui, Hanlin Tang, and SueYeon Chung. 2021.
\newblock \href {https://arxiv.org/abs/2105.14602} {On the geometry of generalization and memorization in deep neural networks}.
\newblock \emph{Preprint}, arXiv:2105.14602.

\bibitem[{von Oswald et~al.(2023)von Oswald, Niklasson, Randazzo, Sacramento, Mordvintsev, Zhmoginov, and Vladymyrov}]{vonoswald2023transformerslearnincontextgradient}
Johannes von Oswald, Eyvind Niklasson, Ettore Randazzo, João Sacramento, Alexander Mordvintsev, Andrey Zhmoginov, and Max Vladymyrov. 2023.
\newblock \href {https://arxiv.org/abs/2212.07677} {Transformers learn in-context by gradient descent}.
\newblock \emph{Preprint}, arXiv:2212.07677.

\bibitem[{Wakhloo et~al.(2023)Wakhloo, Sussman, and Chung}]{wakhloo2023linear}
Albert~J Wakhloo, Tamara~J Sussman, and SueYeon Chung. 2023.
\newblock Linear classification of neural manifolds with correlated variability.
\newblock \emph{Physical Review Letters}.

\bibitem[{Wang et~al.(2024)Wang, Zhu, Saxon, Steyvers, and Wang}]{wang2024largelanguagemodelslatent}
Xinyi Wang, Wanrong Zhu, Michael Saxon, Mark Steyvers, and William~Yang Wang. 2024.
\newblock \href {https://arxiv.org/abs/2301.11916} {Large language models are latent variable models: Explaining and finding good demonstrations for in-context learning}.
\newblock \emph{Preprint}, arXiv:2301.11916.

\bibitem[{Wei et~al.(2022)Wei, Bosma, Zhao, Guu, Yu, Lester, Du, Dai, and Le}]{wei2022finetunedlanguagemodelszeroshot}
Jason Wei, Maarten Bosma, Vincent~Y. Zhao, Kelvin Guu, Adams~Wei Yu, Brian Lester, Nan Du, Andrew~M. Dai, and Quoc~V. Le. 2022.
\newblock \href {https://arxiv.org/abs/2109.01652} {Finetuned language models are zero-shot learners}.
\newblock \emph{Preprint}, arXiv:2109.01652.

\bibitem[{Yerxa et~al.(2023)Yerxa, Kuang, Simoncelli, and Chung}]{yerxa2023learning}
Thomas~Edward Yerxa, Yilun Kuang, Eero~P Simoncelli, and SueYeon Chung. 2023.
\newblock \href {https://openreview.net/forum?id=og9V7NgOrQ} {Learning efficient coding of natural images with maximum manifold capacity representations}.
\newblock In \emph{Thirty-seventh Conference on Neural Information Processing Systems}.

\bibitem[{Zhang et~al.(2015)Zhang, Zhao, and LeCun}]{Zhang2015CharacterlevelCN}
Xiang Zhang, Junbo~Jake Zhao, and Yann LeCun. 2015.
\newblock Character-level convolutional networks for text classification.
\newblock In \emph{NIPS}.

\bibitem[{Zhao et~al.(2021)Zhao, Wallace, Feng, Klein, and Singh}]{zhao2021calibrateuseimprovingfewshot}
Tony~Z. Zhao, Eric Wallace, Shi Feng, Dan Klein, and Sameer Singh. 2021.
\newblock \href {https://arxiv.org/abs/2102.09690} {Calibrate before use: Improving few-shot performance of language models}.
\newblock \emph{Preprint}, arXiv:2102.09690.

\end{thebibliography}

\appendix

\newcolumntype{P}[1]{>{\centering\arraybackslash}p{#1}}

\section{Appendix}
\label{sec:appendix}

\subsection{Dataset details}
\label{sec:appendix_dataset_details}

\subsubsection{Multi-task dataset}

\paragraph{Description}
The synthetic dataset used in this study was generated using Claude 3.5 Sonnet, a large language model developed by Anthropic. The dataset consists of sentences crafted to represent various combinations of emotions, topics, and pragmatic intents. The generation process was designed to create a diverse and balanced dataset suitable for studying representation changes in a multi-task setup. Example sentences covering all five category labels for each of the three subtasks can be found in \cref{tab:multitask_example_sentences}.

\begin{table*}[ht]
    \centering
    \begin{tabular}{P{0.45\linewidth} | P{0.13\linewidth} | P{0.15\linewidth} | P{0.15\linewidth}}
      \textbf{Sentence}  & \textbf{Sentiment} & \textbf{Topic} & \textbf{Intent} \\ \hline
      
    This concert has me over the moon - I'm having the time of my life! & Joy & Entertainment & Idiomatic \\ \hline 

    You'll watch as the winds of change blow away the last remnants of your faith in the system! & Sadness & Politics & Metaphorical \\ \hline 

   Tomorrow's blood test results might reveal something I'm not prepared to handle. & Fear & Health & Literal \\ \hline  

   My favorite team's strategy of constantly fumbling the ball was clearly the path to victory. & Anger & Sports & Sarcastic \\ \hline  

   I nearly fell out of my chair when my ancient printer suddenly sprang to life and started spewing out pages of binary code! & Surprise & Technology & Humorous \\ \hline
   
    \end{tabular}
    \caption{Example sentences from the synthetic multi-task dataset }
    \label{tab:multitask_example_sentences}
\end{table*}

\begin{figure*}
    \centering
    
\begin{tcolorbox}[  enhanced,
  interior style={
    top color=gray!5,
    bottom color=gray!5,
  },
  frame style={
    color=gray!35,
  },
  left=2pt,    % left padding
  right=2pt,   % right padding
  top=2pt,     % top padding
  bottom=2pt,  % bottom padding
]

You are a helpful assistant tasked with generating a dataset of sentences. Generate 2 sentences for each of the following categories of emotion:

1. Joy

2. Sadness

3. Anger

4. Fear

5. Surprise

\medskip

Please make sure all sentences are related to the topic ``\textbf{technology}'' and have \textbf{humorous} intent (\textbf{Each sentence is intended to be funny or amusing, often through clever use of language, unexpected connections, or playful exaggeration.}).

\medskip
There are a few requirements for the sentences:

Use \textbf{first-person} perspective.

Use \textbf{future} tense.

\medskip
Additionally, include at least one sentence that:

- \textbf{Sounds like a tweet}

\medskip

Very important instructions:

1. Convey the emotion through the situation, word choice, and tone. Do not directly state the emotion or use immediate synonyms.

2. Imply the topic through context and content, but do not explicitly mention the topic name.

3. Express the intent naturally without explicitly stating the type of intent being used.

\medskip

Format your response as follows:

Joy:

1. [Sentence 1]

2. [Sentence 2]

\smallskip
Sadness:

1. [Sentence 1]

2. [Sentence 2]

\smallskip
Anger:

1. [Sentence 1]

2. [Sentence 2]

\smallskip
Fear:

1. [Sentence 1]

2. [Sentence 2]

\smallskip

Surprise:

1. [Sentence 1]

2. [Sentence 2]

\medskip
Ensure each sentence is on a new line and numbered within its category.

Do not include any additional text or explanations outside of this format.

Very important: Remember to vary the syntax and structure of the sentences to make the dataset diverse and interesting! Do not use the same structure for all sentences.

\end{tcolorbox}
    \caption{Example prompt configuration used in generating the synthetic dataset (emotion-focused type). Text highlighted in \textbf{bold} represents parts of the prompt that were varied on each iteration to increase diversity of resulting sentences.}
    \label{fig:example-generation-prompt}
\end{figure*}

\paragraph{Generation procedure}

The dataset was generated through an iterative process, by cycling through three possible generation types:

\begin{itemize}
\item Emotion-focused: The model's goal was to generate 10 sentences (2 for each emotion), given a specific topic and intent.
\item Topic-focused: The model's goal was to generate 10 sentences (2 for each topic), given a specific emotion and intent.
\item Intent-focused: The model's goal was to generate 10 sentences (2 for each intent), given a specific emotion and topic.
\end{itemize}

Example of a full prompt an for emotion-focused iteration can be found in \cref{fig:example-generation-prompt}. Prompts for other two types were phrased analogously.

To ensure diversity of resulting sentences, each prompt included a specific constraint for the sentence to be of a certain perspective (first, second or third person) and tense (present, past, future or mixed), chosen at random.

Additionally, the model was instructed to make at least one generated sentence follow a special requirement, that was sampled randomly from a pool of 18 possible requirements (\cref{tab:special-requirements}) during prompt construction. 

\begin{table*}[ht]
    \centering
    \begin{tabular}{p{0.75\linewidth}}
        Sounds like a tweet \\
        Describes a hypothetical scenario \\
        Uses simple vocabulary as if spoken by a child \\
        Has a rhythmic or lyrical quality \\
        Sounds like a memorable quote \\
        Includes a question \\
        Includes a command or instruction \\
        Incorporates a well-known saying or proverb \\
        Structured like a headline \\
        Includes a number or statistic \\
        Imitates casual online comment style \\
        Uses formal language \\
        Starts with a gerund (-ing word) \\
        Includes a rhetorical question \\
        Uses the passive voice \\
        Includes a list or enumeration \\
        Employs repetition for emphasis \\
        Starts with a conditional (If...)
    \end{tabular}
    \caption{Possible special requirements during dataset generation}
    \label{tab:special-requirements}
\end{table*}

The generated sentences were parsed and stored with their corresponding labels.

\paragraph{Post processing}

After generation, the dataset underwent several post-processing steps:

\begin{itemize}
    \item Duplicate sentences were removed to ensure uniqueness
    \item Category labels were capitalized and ordered consistently.
    \item Letter codes and shuffled labels were assigned to each sentence for alternative labeling schemes
    \item Dataset was subsampled to 1000 sentences (500 for train and 500 for test splits) in a way to ensure uniform coverage of each of three subtasks (in each split: 100 sentences per category)
\end{itemize}

To introduce further variations, the following transformations were applied to both train and test sets:
\begin{itemize}
\item 10\% of sentences were converted to lowercase
\item 10\% of sentences were converted to uppercase
\end{itemize}
These transformations were applied randomly and independently to each set.

\subsubsection{Open datasets}

To ensure our findings were not solely a result of using a synthetic dataset generated by another language model, we replicated our single-task experiments using two open datasets, often used for text classification: \textbf{AG News} \cite{Zhang2015CharacterlevelCN}and \textbf{TREC (coarse)} \cite{li-roth-2002-learning, hovy-etal-2001-toward}. For the TREC dataset we removed the ``Abbreviation'' category, which had an insufficient number of samples for manifold analysis. Additionally, we created a balanced test partition with uniform representation across all categories. Resulting dataset sizes can be found in \cref{tab:open_datasets_samples}.

\begin{table*}[ht]
    \centering
    \begin{tabular}{c|c|c}
        \textbf{Dataset} & \textbf{Samples per category} & \textbf{Category labels} \\  \hline
        TREC coarse & 65 & Description, Entity, Human, Location, Numeric \\  \hline
        
        AG news & 63 & Business, World, Sports, Sci/Tech \\  \hline
    \end{tabular}
    \caption{Parameters of open dataset subsampling sizes used in experiments}
    \label{tab:open_datasets_samples}
\end{table*}

\subsection{Models}
All experiments presented in the main text were performed on Llama3.1 8b base model \cite{dubey2024llama3herdmodels} (32 layers, 4096 embedding dimension). We also repeated the results with Gemma2 (2b base model) \cite{gemmateam2024gemma2improvingopen} (26 layers, 2304 embedding dimension). Results are presented in the \cref{sec:supplementary_plots}.

\subsection{Methods}

\subsubsection{Embedding Extraction Methodology}
\label{appendix:embedding_extraction}

\paragraph{Challenges in Decoder-Only Models}

\begin{enumerate}
    \item \textbf{Masked Self-Attention:} In decoder-only models, each token's embedding is limited to information from itself and preceding tokens. This requires the model to progressively accumulate and propagate relevant contextual information along the sequence, influencing how global features are represented across different token positions.

    \item \textbf{Distributed Sentence-Level Features:} Unlike models with dedicated [CLS] tokens, global sentence-level features (such as sentiment) might be distributed across embedding vectors of intermediate tokens.

    \item \textbf{Last Token Dependency:} The model's output is a function of the last token's embedding vector only, implying that task-relevant features must be aggregated and represented in this final embedding for good task performance.

\end{enumerate}

\paragraph{Sentence Embeddings}
To examine how task-specific prompts influence feature extraction and computation on intermediate tokens, we construct sentence embeddings as follows:

\begin{enumerate}
    \item We extract residual stream activations at each layer for tokens corresponding only to the input sentence, excluding the task prompt itself. This ensures that the resulting embeddings for each sentence are of the same length across different prompting conditions.

    \item We perform mean-pooling across these embedding vectors to obtain a fixed-size embedding for each sentence.
\end{enumerate}

While this method differs from using dedicated sentence-level embeddings, it provides insight into the model's intermediate processing stage. Based on the idea of feature superposition, we hypothesize that directions in the embedding space corresponding to task-irrelevant token-level features will be averaged out, while task-relevant global features (which might be distributed among various tokens in the sentence) will be preserved or enhanced through mean-pooling.

\paragraph{Last-token Embeddings}
While mean-pooled embeddings allow us to capture an intermediate processing stage, the underlying sentence tokens are not immediately utilized for the task. To understand the model's final representation before output generation, we also extracted residual stream activation of the last token in the sequence at each layer. The last token is special because, for the model to perform the task, all relevant sentence-level features must get "packaged" into the embedding vector of the last token via self-attention. By analyzing last token embeddings across layers, we can track at what point such feature repackaging takes place to collect information about the sentence.

\subsubsection{Manifold Capacity}
\label{appendix:manifold_capacity}

This section provides additional background on the idea of manifold capacity. Consider a set of $N$ points in $D_\text{emb}$-dimensional space: $\vec x_i \in \mathbb R^{D_\text{emb}}$. Each point has a corresponding class label $l_i \in \{1,\dots P\ \}$. Capacity measures how well a particular representation supports linear separability of a random one-vs-rest label dichotomy that doesn't break category boundaries. Namely, for $P$ classes there are $P$ possible dichotomies: $\{ y_i^\mu \}$, where $i \in \{1,\dots N\ \}$ – index of a data point, $\mu \in \{1,\dots P\ \}$ – index of a dichotomy, and:
$$
\begin{cases} y_i^\mu = 1 \ \operatorname{if} \left( l_i = \mu \right)  \\
y_i^\mu = -1 \ \operatorname{otherwise}
\end{cases}
$$
Consider performing a random projection of data into a $D_\text{proj}$- dimensional space, where $D_\text{proj} \leq D_\text{emb}$. We can compute a probability that a randomly chosen dichotomy will be linearly separable, when the data is projected randomly to $D_\text{proj}$ dimensions, formalized as follows:

$$
F(D_\text{proj}) = \displaystyle \mathop{\operatorname{Pr}}_{\substack{
S\sim \mathcal N^{(D_\text{proj}, D_\text{emb})} \\
                  \mu \sim \text{Unif.}(\{1 \dots P\})}
 \\ } \left[  \exists \vec w: \ y_i^\mu \vec w S \vec x_i \geq 0 \ \forall i 
\right]
$$

Where $\vec w\in D_\text{proj}$. In a thermodynamic limit of $N, P \to \infty$, $F(D_\text{proj}$) undergoes a sharp phase transition from $0$ to $1$ as $D_\text{proj}$ interpolates between $0$ and $D_\text{emb}$. In the finite data case, the transition is smooth, but we can still detect an approximate critical dimension $D^*$, that corresponds to the inflection point of $F(D_\text{proj})$. Then, manifold capacity $\alpha$ is defined to be
$$
\alpha = \frac{P}{D^*}
$$

Intuitively it captures decoding efficiency, quantifying how many dimensions are sufficient for a downstream readout to perform classification. $\alpha$ depends on the geometry of individual manifolds (such as radius and dimension), as well as relative positioning and alignment of different class manifolds in the embedding space.

\subsubsection{Manifold dimension}

We use the participation ratio (PR) as a proxy for manifold dimensionality, as described in \cite{gaoTheoryMultineuronalDimensionality2017}. For a manifold $\mathbf{X} \in \mathbb{R}^{(N, D)}$ consisting of $N$ points in a $D$-dimensional space ($N\leq D$), the participation ratio is defined as:
$$
\operatorname{PR} = \frac{\left( \sum_i^N \lambda_i  \right)^2}{\sum_i^N \lambda_i^2 }
$$
where $\lambda_i$ is the $i^\text{th}$ eigenvalue of the manifold covariance matrix $\mathbf{X} \mathbf{X}^T$.
Intuitively, PR measures how evenly the total variance is distributed among individual principal components. Lower values of $\operatorname{PR}$ indicate a more rapid decay of covariance eigenvalues, signifying lower effective dimensionality. We compute the PR for each manifold and then average these values to obtain a single measure of dimensionality for the entire representation.
 
\subsubsection{In-context learning}

\paragraph{Demonstration Prompts}
We constructed demonstration prompts by randomly sampling sentences from the training split. The number of examples varied from 1 to 40, ensuring as uniform a label coverage as possible. For instance, in a 4-category classification task with 10 demonstration examples, 8 examples were guaranteed to cover all categories equally (2 per category), with the remaining 2 examples randomly chosen. We computed the forward pass of the model with 3 random seeds for each number of demonstrations and reported averaged measures across these runs.

\paragraph{Accuracy Measurement}
We measured accuracy as the proportion of test sentences for which the token with the highest logit value corresponded to the first token of the target output (for cases where the target label was tokenized into multiple tokens). Importantly, we considered logits for the entire vocabulary, not restricting the scope to target outputs. If the highest probability output was a token not corresponding to any class label, the run was treated as incorrect, irrespective of relative logit values of other tokens.

\subsubsection{Prompt-tuning}

\paragraph{Description}
We replicated our main experimental setup, replacing natural language task instructions and demonstrations with tunable prompts of varying lengths. A tunable prompt $\mathbf{X}$ (also referred to as a soft prompt) of length $l$ is a matrix in the model's embedding space $\mathbb{R}^{l \times D_\text{emb}}$, where $D_\text{emb}$ is the dimensionality of the model's token embeddings.

Unlike discrete text prompts, these tunable prompts are continuous vectors that can be optimized directly through gradient descent. They provide a more flexible way to convey task-specific information to the model, unconstrained by the token embedding matrix. This allows them to occupy highly specific regions of the embedding space that are inaccessible through natural language input alone.

\paragraph{Soft-prompt methodology}

\begin{enumerate}
    \item \textbf{Initialization}: Each tunable prompt $\mathbf{X}$ was initialized using the embedding vector of the word ``Category''. For soft prompts with $l > 1$, this embedding vector was repeated $l$ times along the sequence length dimension, providing a starting point for optimization.

    \item \textbf{Prepending}: For each input sequence $\mathbf{s}$ (after token embedding), we prepended the tunable prompt $\mathbf{X}$ to create an augmented input:
    
    \[ \mathbf{s}_\text{augmented} = [\mathbf{X}; \mathbf{s}] \]
    
    where $[;]$ denotes concatenation along the sequence length dimension.

    \item \textbf{Optimization}: During training, while keeping the pretrained language model parameters fixed, we optimized the elements of $\mathbf{X}$ to minimize the task-specific loss function:
    
    \[ \mathbf{X}^* = \text{argmin}_\mathbf{X} \mathcal{L}(\text{Model}([\mathbf{X}; \mathbf{s}]), \mathbf{y}) \]
    
    where $\mathcal{L}$ is the Cross Entropy loss function, $\text{Model}(\cdot)$ represents the frozen pretrained language model, and $\mathbf{y}$ is the ground truth label.

    \item \textbf{Length Variation}: We trained soft prompts of various lengths $l \in \{1,2,5,10,20\}$ to investigate the impact of prompt size on performance. Longer prompts can theoretically capture more details about general task structure, the nature of categories, and meta-information about specific training examples (although in practice, we did not observe significant performance differences across different lengths).
    
\end{enumerate}

\paragraph{Training procedure and checkpoints}

Soft prompts were optimized on the training subset of each dataset (see \cref{sec:appendix_dataset_details}). We trained each soft prompt for 30 epochs with a batch size of 16 using the Adam optimizer \cite{kingma2017adammethodstochasticoptimization}. The initial learning rate was set to $3 \times 10^{-4}$ with an exponential decay of $\gamma = 0.9$ after each epoch.
To analyze how representations evolved during the training of soft prompts, we selected 50 intermediate points, logarithmically spaced across training iterations.

\subsection{Computational resources}
All experiments were performed on a high-performance computing cluster, using Nvidia H100 GPUs, resulting in total of 1000 GPU hours.

\newpage

\subsection{Supplementary plots}

To maintain a reasonable number of figures in the paper, we present a curated subset in this appendix, highlighting key points with representative plots. The complete set of figures, detailing geometric measures for all combinations of models, datasets, and tasks, along with the source code, will be available on GitHub. The repository will be made public upon publication.

\label{sec:supplementary_plots}

\begin{figure*}
    \centering
    \includegraphics[width=1\linewidth]{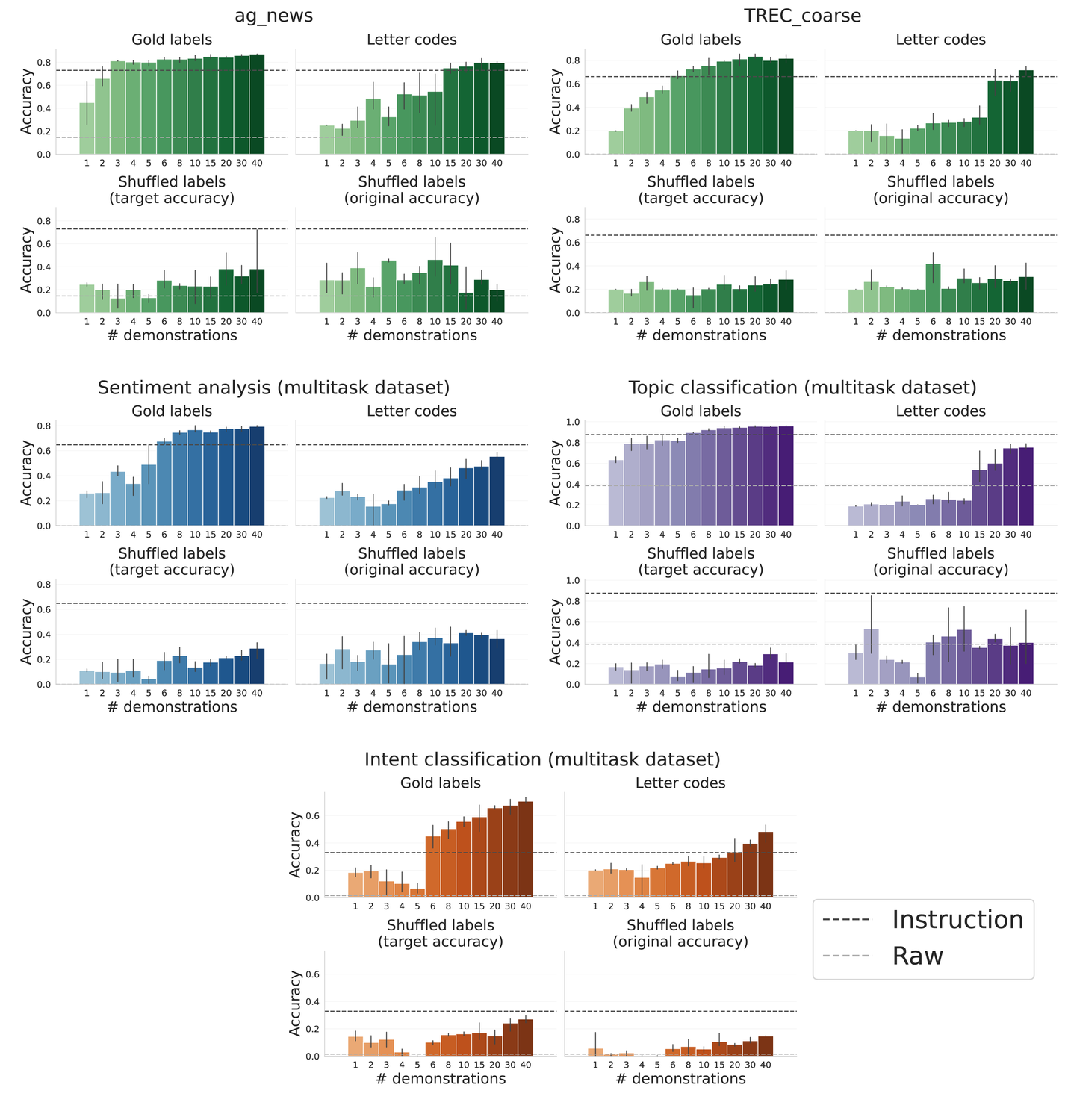}
    \caption{\textbf{Llama3.1-8b} performance of demonstrations and instruction prompts on open datasets (ag\_news and TREC coarse) and on all three subtasks of the synthetically generated multitask dataset (sentiment, topic and intent).}
    \label{fig:ICL_extended_performance_llama}
\end{figure*}

\begin{figure*}
    \centering
    \includegraphics[width=1\linewidth]{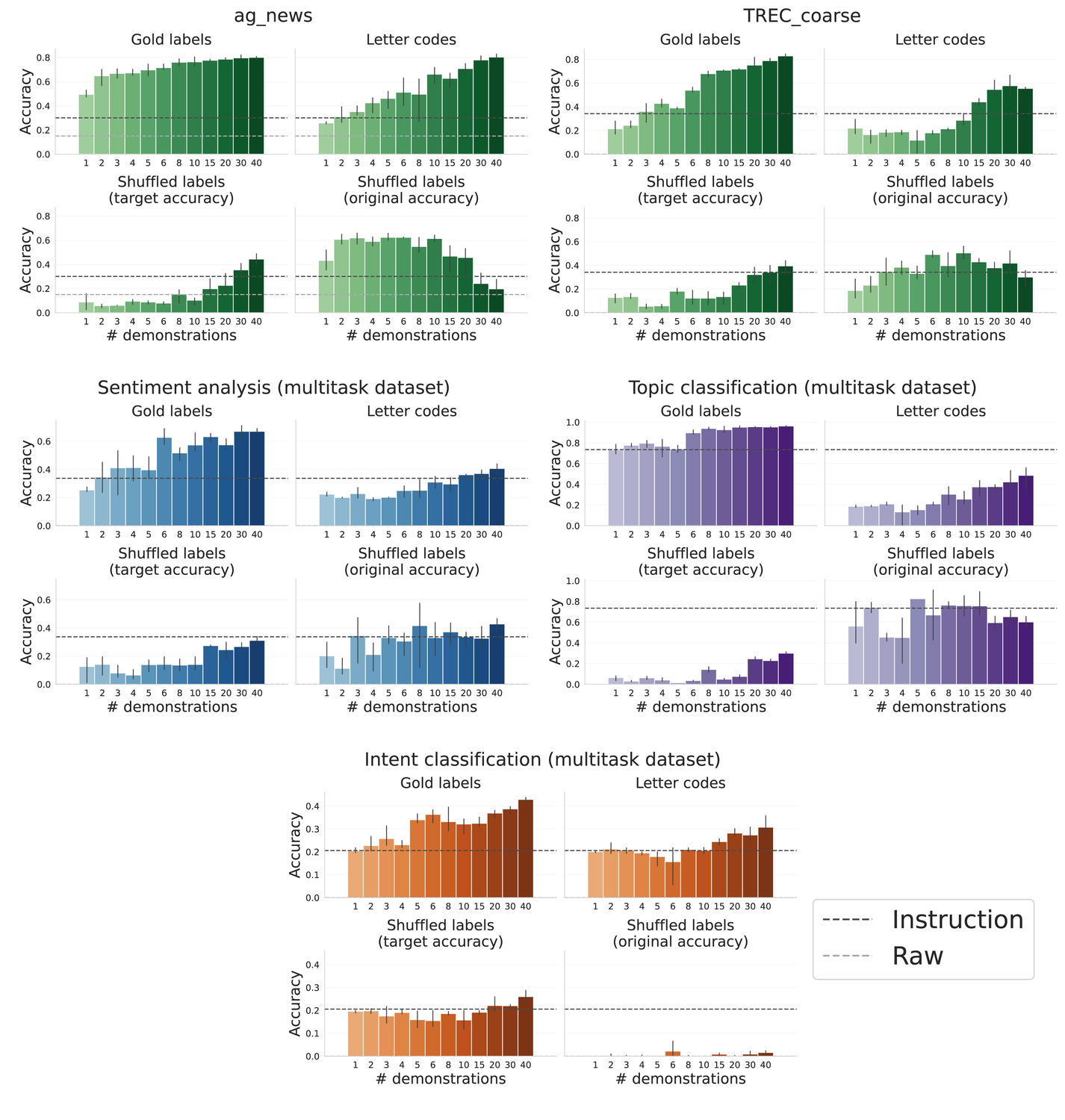}
    \caption{\textbf{Gemma2-2b} performance of demonstrations and instruction prompts on open datasets (ag\_news and TREC coarse) and on all three subtasks of the synthetically generated multitask dataset (sentiment, topic and intent).}
    \label{fig:ICL_extended_performance_gemma}
\end{figure*}

\begin{figure*}
    \centering
    \includegraphics[width=0.85\linewidth]{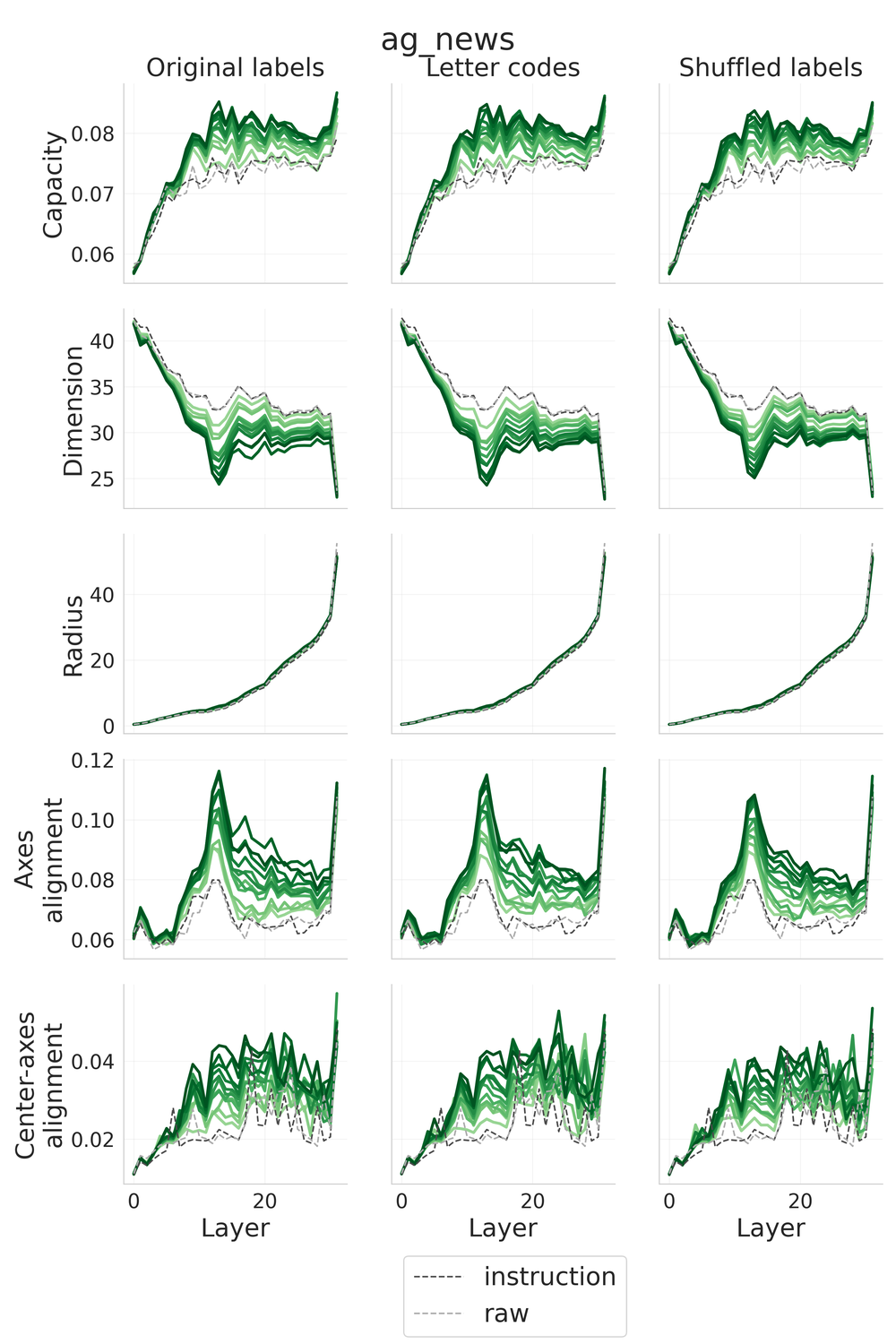}
    \caption{Manifold capacity and geometric properties of \textbf{sentence-level} representations during demonstration prompting compared to instruction and raw sentence across layers. \textbf{Llama3.1-8b} evaluated on \textbf{ag\_news}. Gradient color shows number of demonstrations (darker --- more examples).}
\end{figure*}

\begin{figure*}
    \centering
    \includegraphics[width=0.85\linewidth]{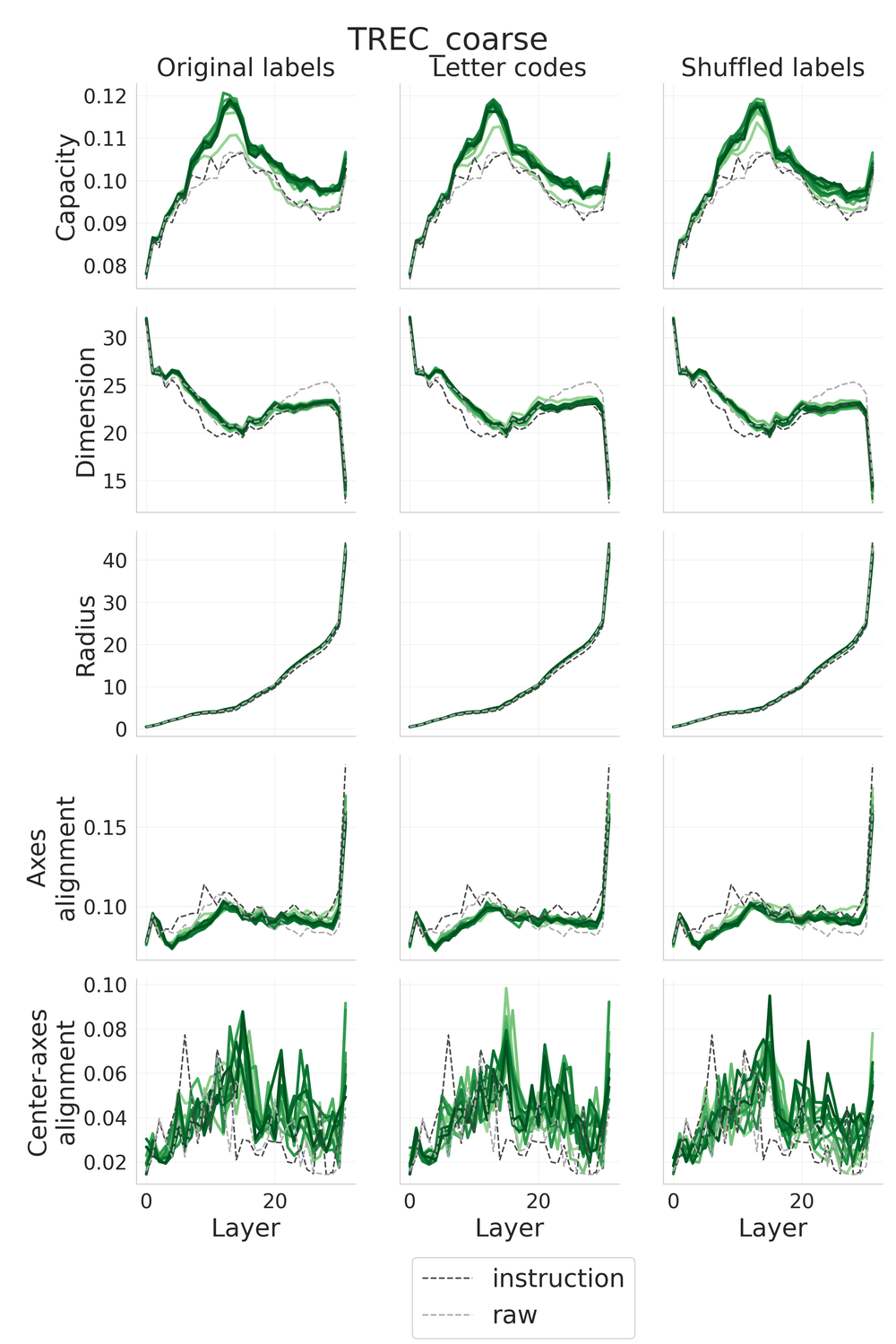}
    \caption{Manifold capacity and geometric properties of \textbf{sentence-level} representations during demonstration prompting compared to instruction and raw sentence across layers. \textbf{Llama3.1-8b} evaluated on \textbf{TREC\_coarse}. Gradient color shows number of demonstrations (darker --- more examples).}
\end{figure*}

\begin{figure*}
    \centering
    \includegraphics[width=0.85\linewidth]{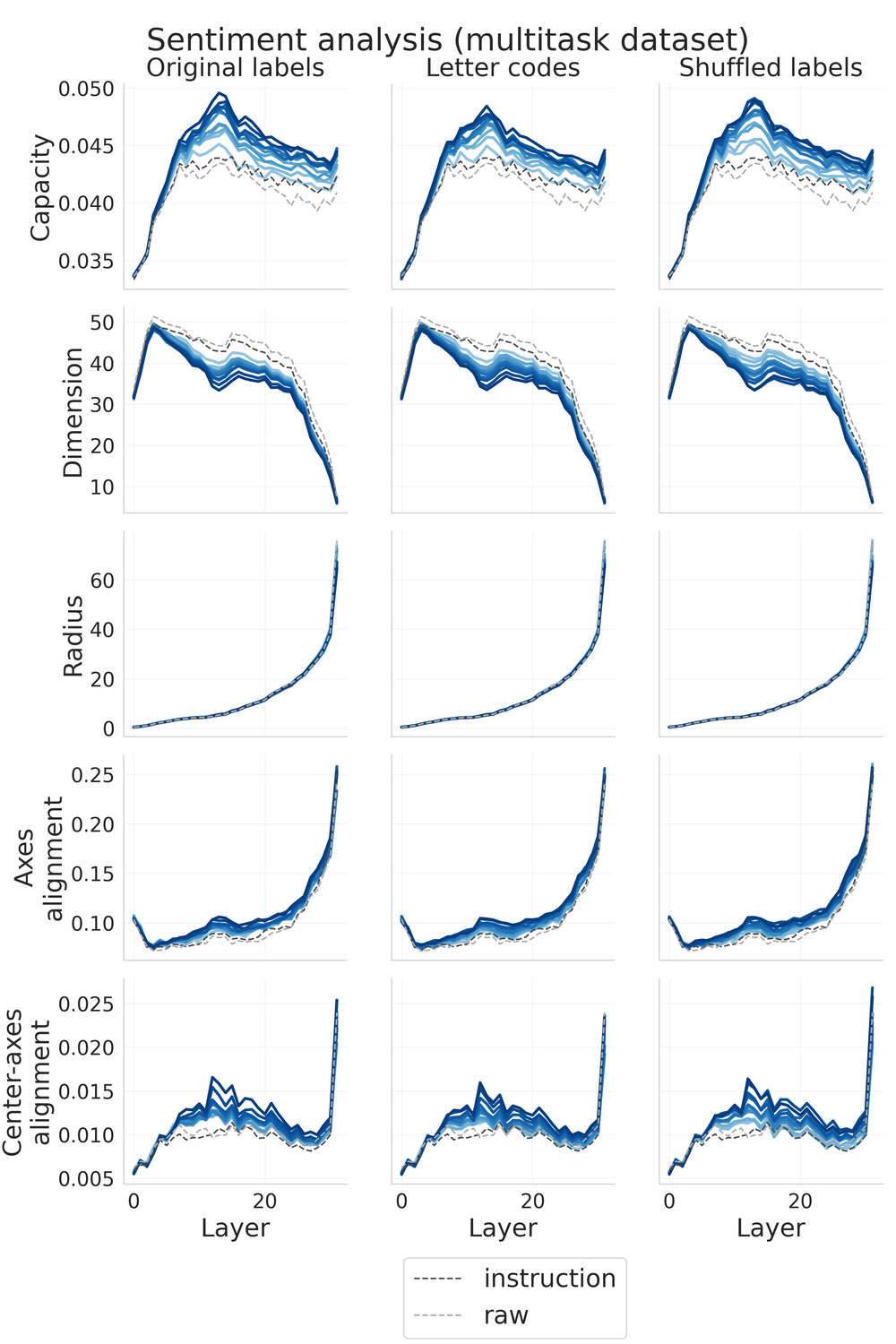}
    \caption{Manifold capacity and geometric properties of \textbf{sentence-level} representations during demonstration prompting compared to instruction and raw sentence across layers. \textbf{Llama3.1-8b} evaluated on \textbf{sentiment analysis} subtask of multitask synthetic dataset. Gradient color shows number of demonstrations (darker --- more examples).}
    \label{fig:ICL_sentence_level_extended}
\end{figure*}

\begin{figure*}
    \centering
    \includegraphics[width=0.85\linewidth]{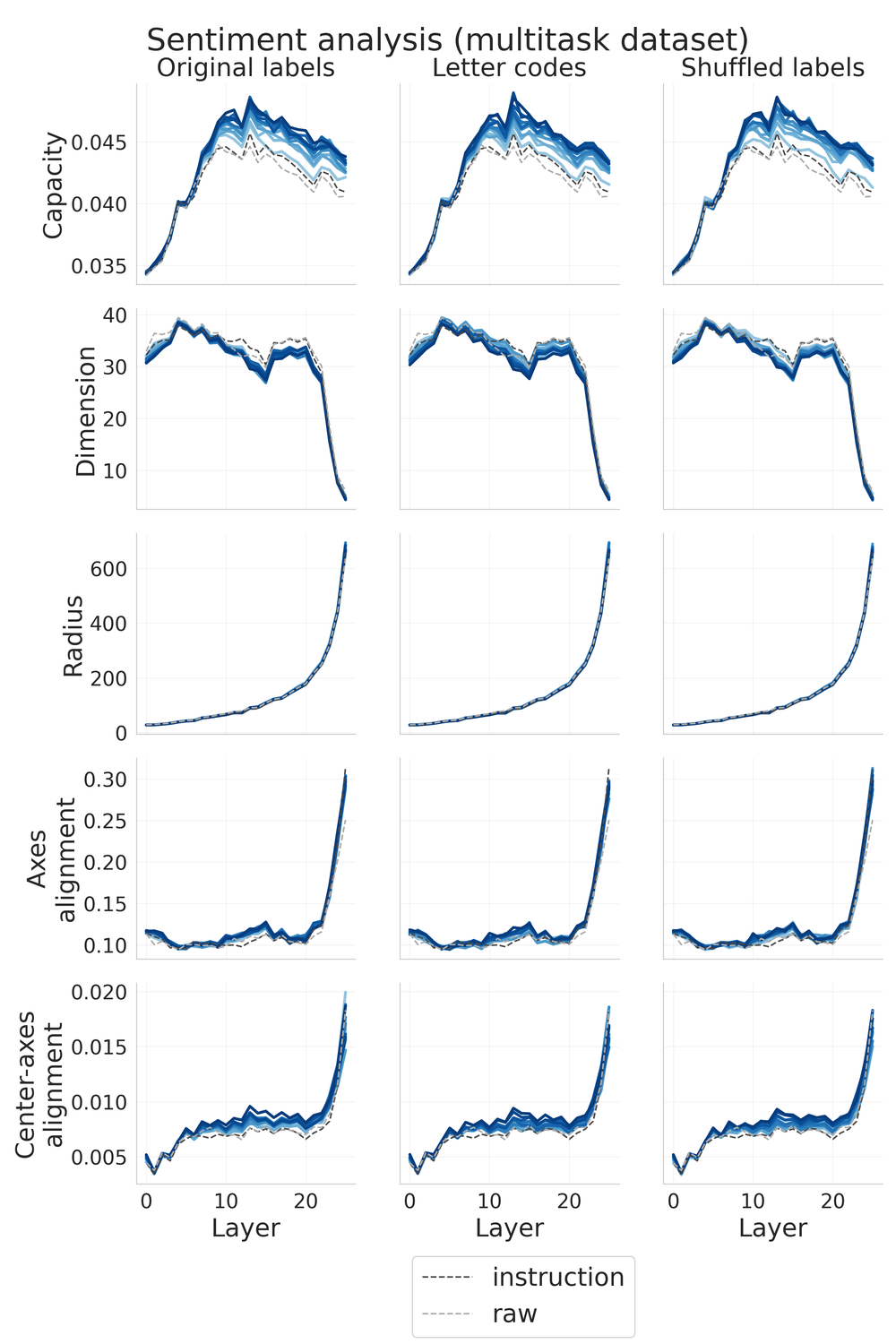}
    \caption{Manifold capacity and geometric properties of \textbf{sentence-level} representations during demonstration prompting compared to instruction and raw sentence across layers. \textbf{Gemma2-2b} evaluated on \textbf{sentiment analysis} subtask of multitask synthetic dataset.Gradient color shows number of demonstrations (darker --- more examples).}
\end{figure*}

% ------------------- LAST TOKEN --------------------------

\begin{figure*}
    \centering
    \includegraphics[width=0.85\linewidth]{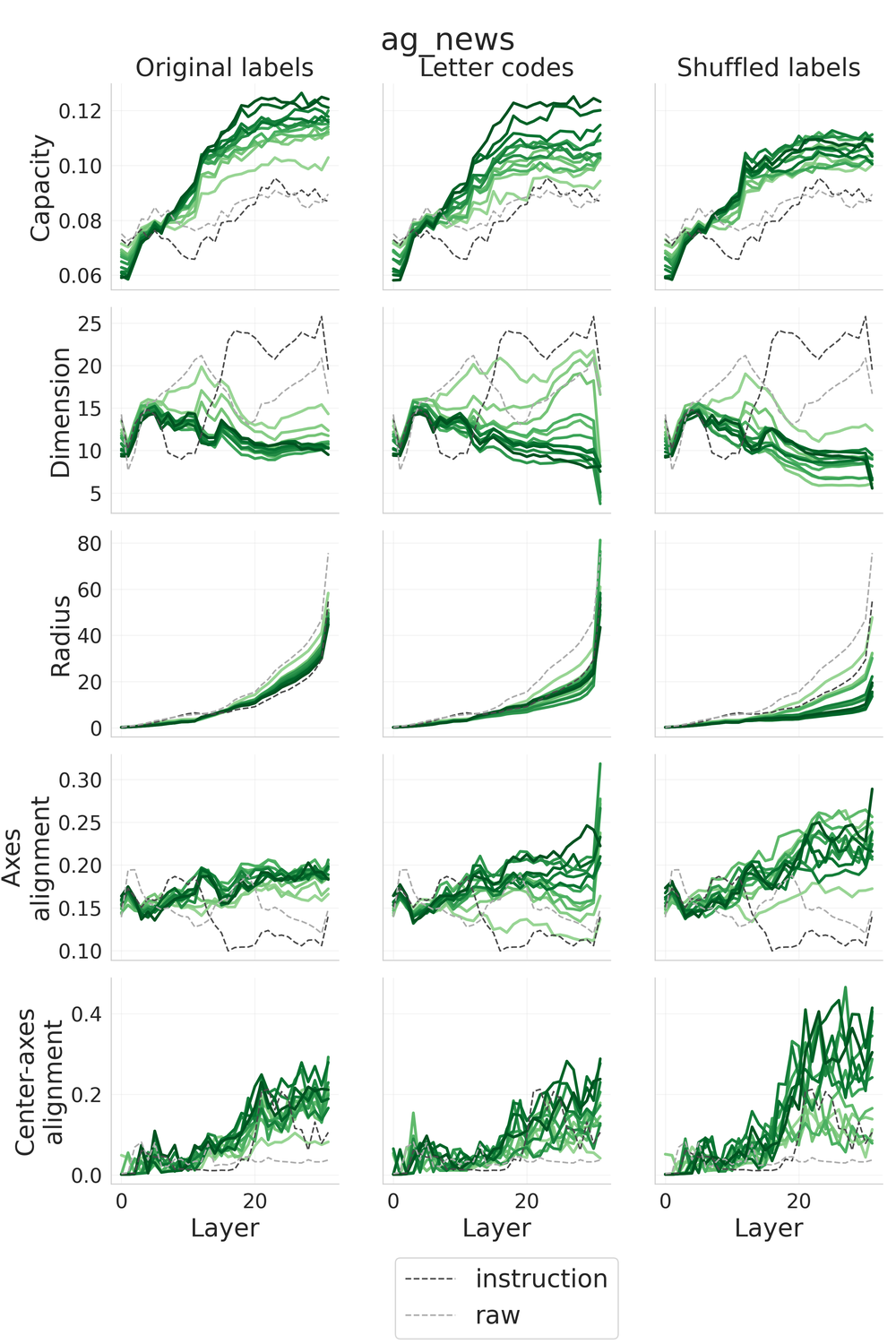}
    \caption{Manifold capacity and geometric properties of \textbf{last token} representations during demonstration prompting compared to instruction and raw sentence across layers. \textbf{Llama3.1-8b} evaluated on \textbf{ag\_news}. Gradient color shows number of demonstrations (darker --- more examples).}
\end{figure*}

\begin{figure*}
    \centering
    \includegraphics[width=0.85\linewidth]{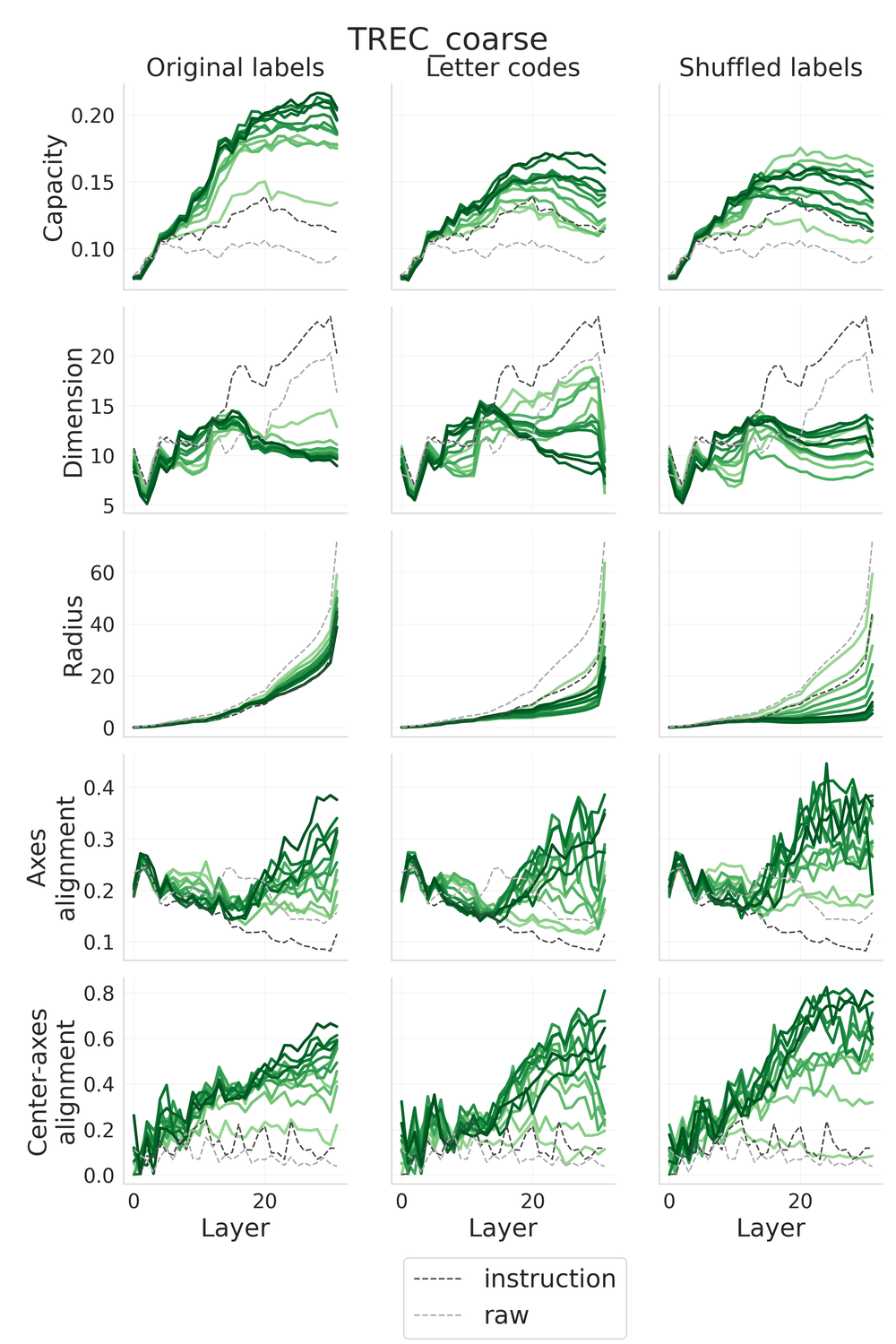}
    \caption{Manifold capacity and geometric properties of \textbf{last token} representations during demonstration prompting compared to instruction and raw sentence across layers. \textbf{Llama3.1-8b} evaluated on \textbf{TREC\_coarse}. Gradient color shows number of demonstrations (darker --- more examples).}
\end{figure*}

\begin{figure*}
    \centering
    \includegraphics[width=0.85\linewidth]{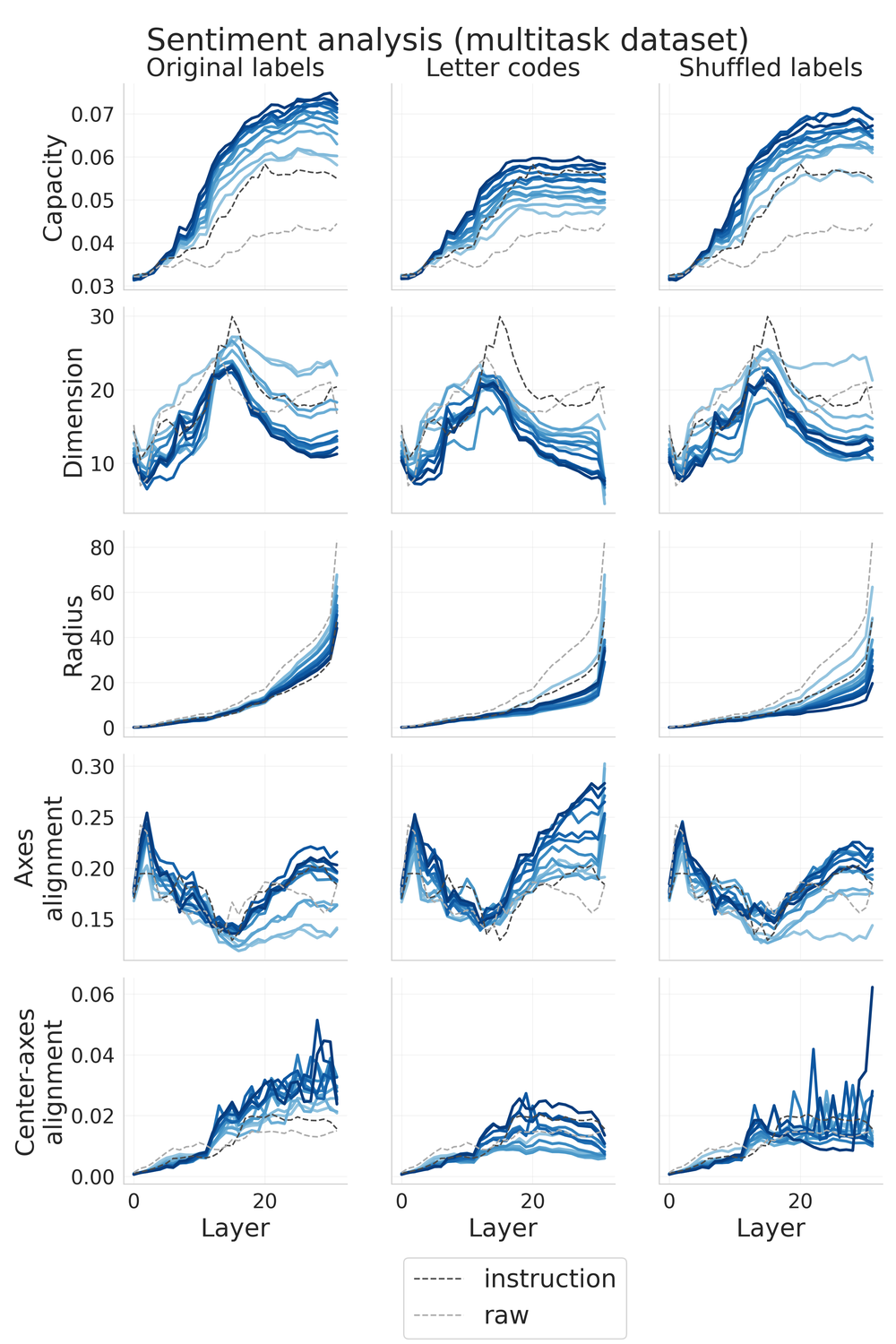}
    \caption{Manifold capacity and geometric properties of \textbf{last token} representations during demonstration prompting compared to instruction and raw sentence across layers. \textbf{Llama3.1-8b} evaluated on \textbf{sentiment analysis} subtask of multitask synthetic dataset. Gradient color shows number of demonstrations (darker --- more examples).}
    \label{fig:ICL_last_token_extended}
\end{figure*}

\begin{figure*}
    \centering
    \includegraphics[width=0.85\linewidth]{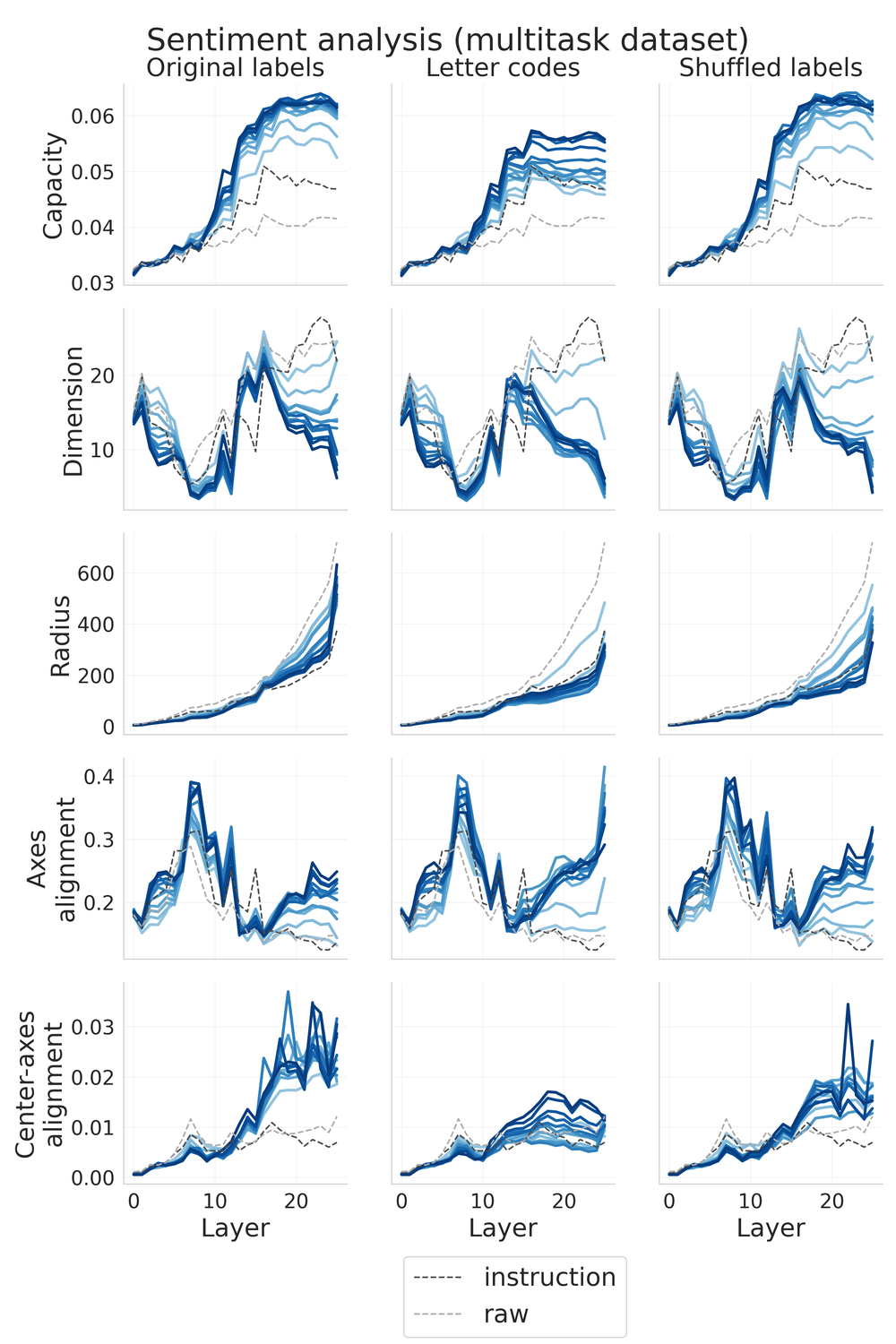}
    \caption{Manifold capacity and geometric properties of \textbf{last token} representations during demonstration prompting compared to instruction and raw sentence across layers. \textbf{Gemma2-2b} evaluated on \textbf{sentiment analysis} subtask of multitask synthetic dataset. Gradient color shows number of demonstrations (darker --- more examples). }
\end{figure*}

\begin{figure*}[h!]
    \centering
    \includegraphics[width=1\linewidth]{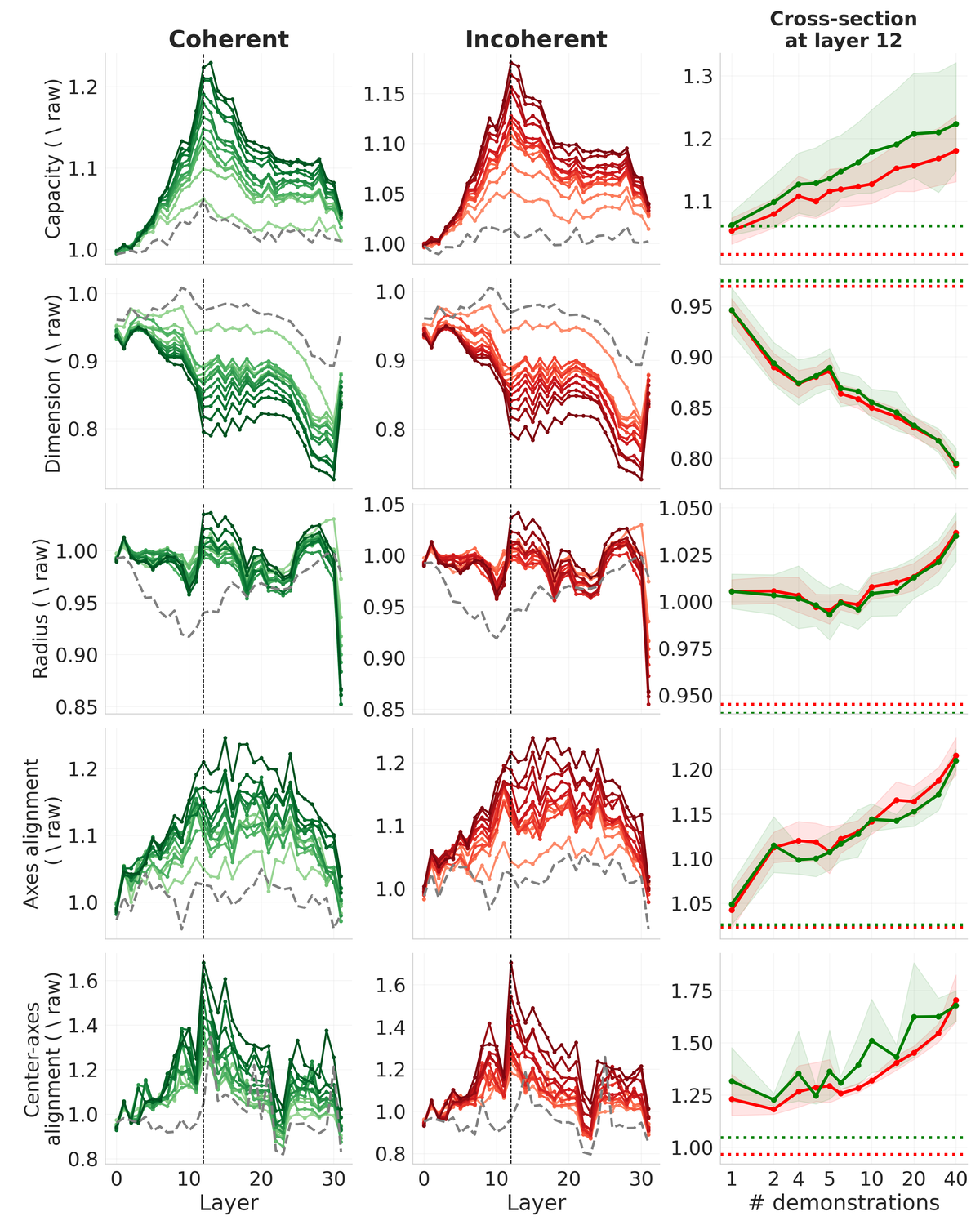}
    \caption{Geometric measures of \textbf{sentence-level} representation during coherent and incoherent task-prompting of \textbf{Llama3.1-8b}.Gradient color shows number of demonstration examples (darker --- more examples). Dashed lines --- instruction prompt.}
    \label{fig:multitask_extended_sentence}
\end{figure*}

\begin{figure*}[h!]
    \centering
    \includegraphics[width=1\linewidth]{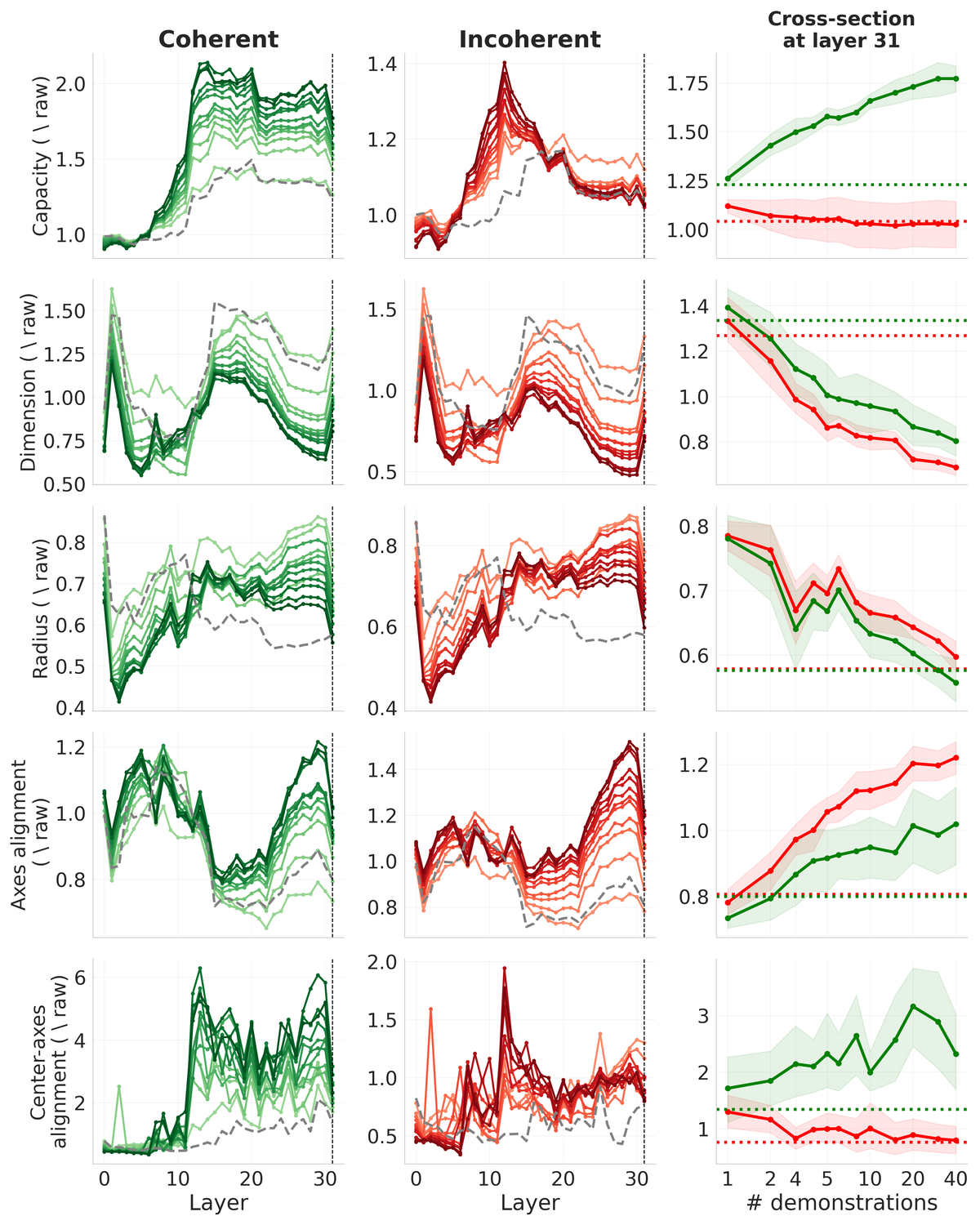}
    \caption{Geometric measures of \textbf{last-token} representation during coherent and incoherent task-prompting of \textbf{Llama3.1-8b}. Gradient color shows number of demonstration examples (darker --- more examples). Dashed lines --- instruction prompt.}
    \label{fig:multitask_extended_last_token}
\end{figure*}

\begin{figure*}[h!]
    \centering
    \includegraphics[width=1\linewidth]{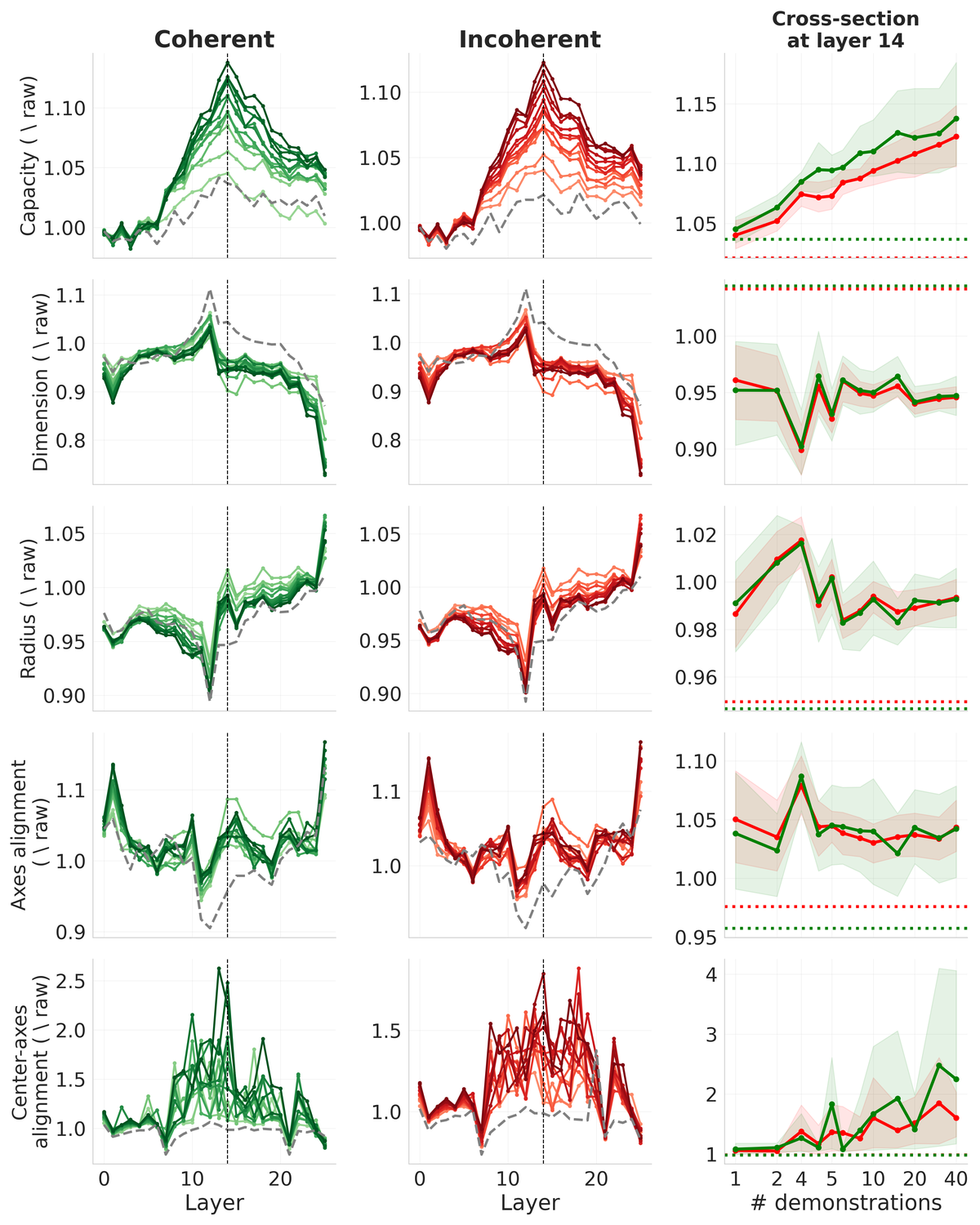}
    \caption{Geometric measures of \textbf{sentence-level} representation during coherent and incoherent task-prompting of \textbf{Gemma2-2b}. Gradient color shows number of demonstration examples (darker --- more examples). Dashed lines --- instruction prompt}
\end{figure*}

\begin{figure*}[h!]
    \centering
    \includegraphics[width=1\linewidth]{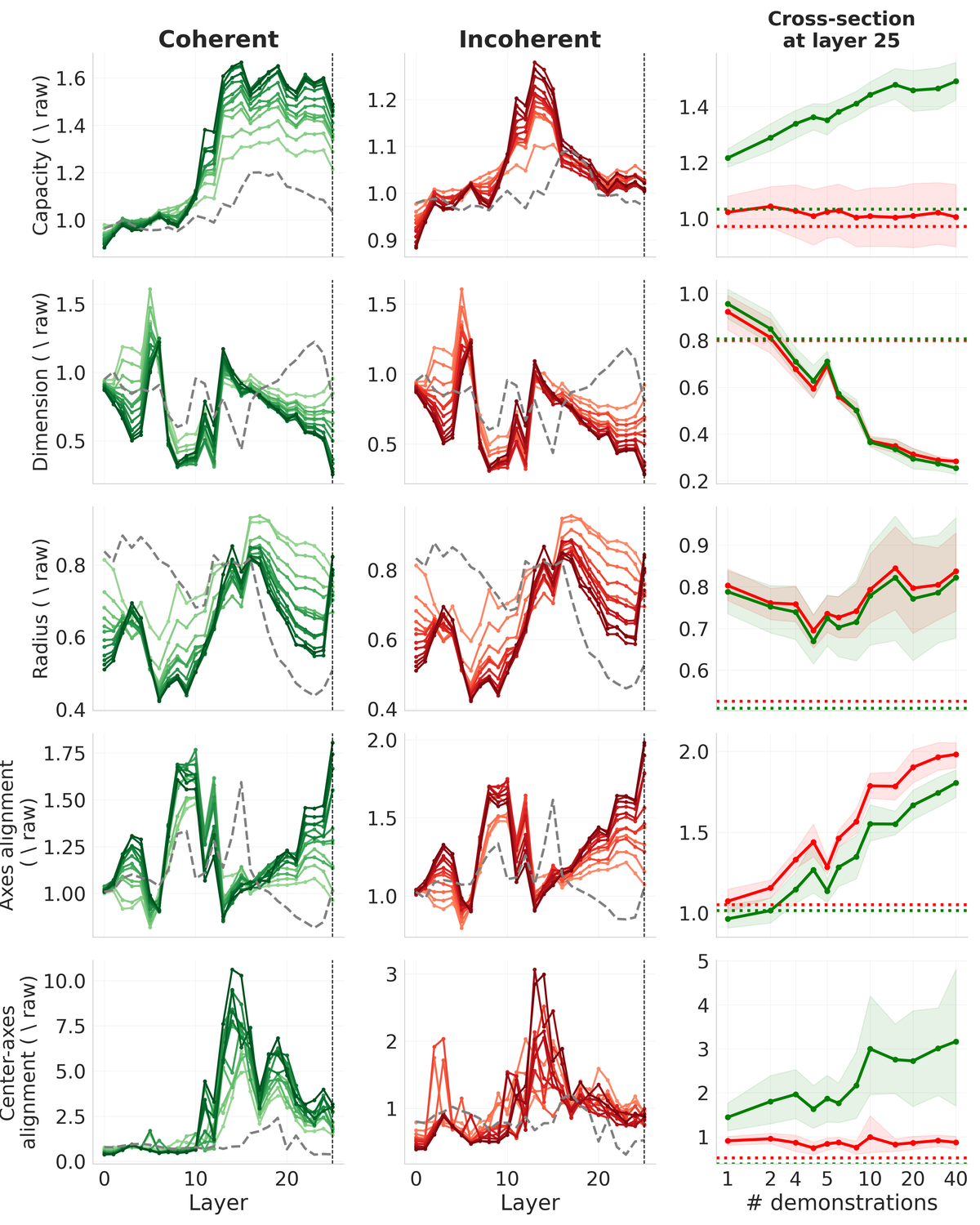}
    \caption{Geometric measures of \textbf{last-token} representation during coherent and incoherent task-prompting of \textbf{Gemma2-2b}. Gradient color shows number of demonstration examples (darker --- more examples). Dashed lines --- instruction prompt.}
\end{figure*}

\label{sec:prompt-tuning-sentence-level}

\begin{figure*}
    \centering
    \includegraphics[width=1\linewidth]{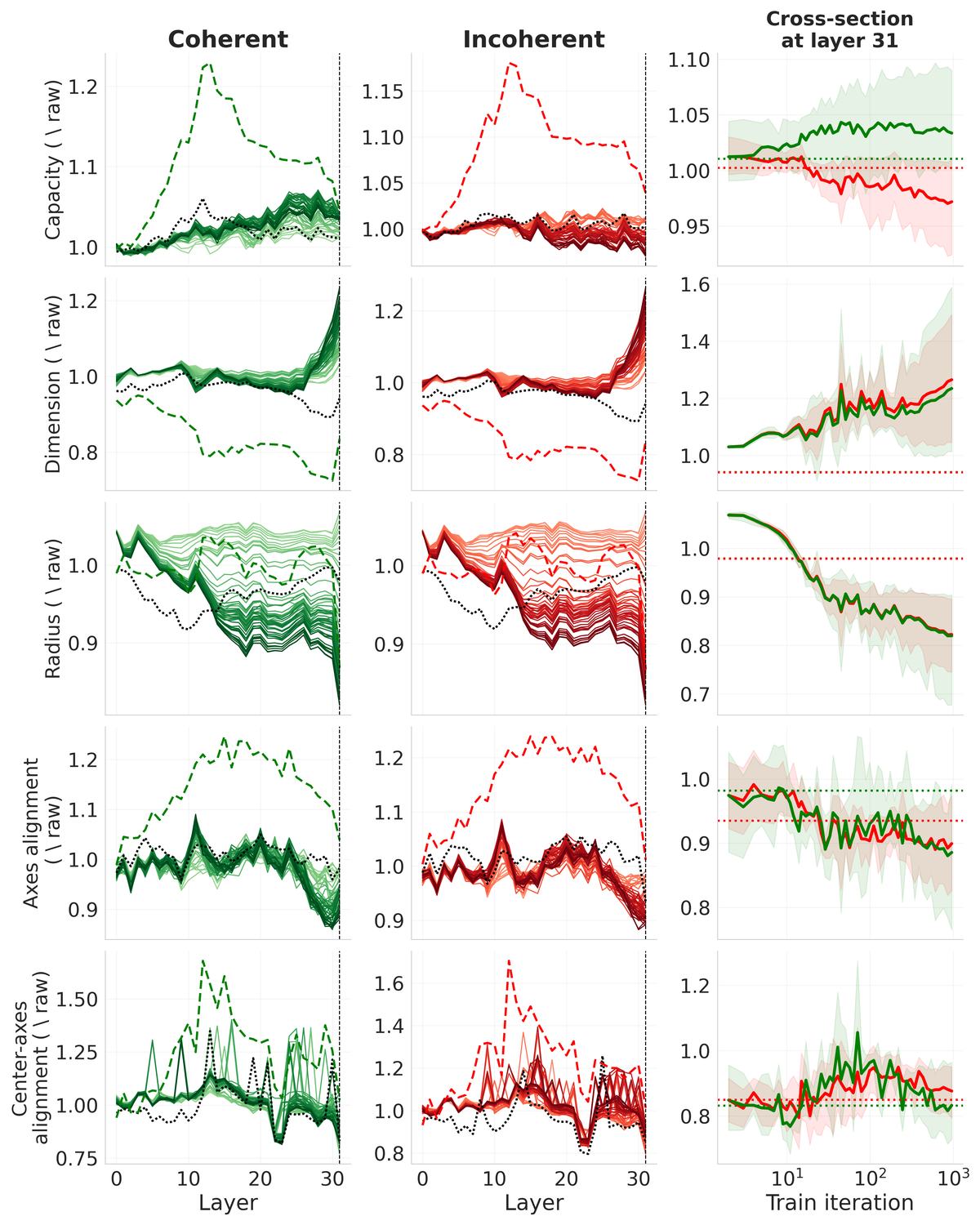}
    \caption{Manifold capacity and geometric measures of \textbf{sentence-level} embeddings during training soft-prompt of length 5 (Llama3.1-8b). Gradient color shows training iterations (darker --- later epochs). Dashed lines --- demonstrations prompt with 40 examples for reference. Dotted --- instruction prompt.}
    \label{fig:prompt_tuning_extended_sentence}
\end{figure*}

\begin{figure*}
    \centering
    \includegraphics[width=1\linewidth]{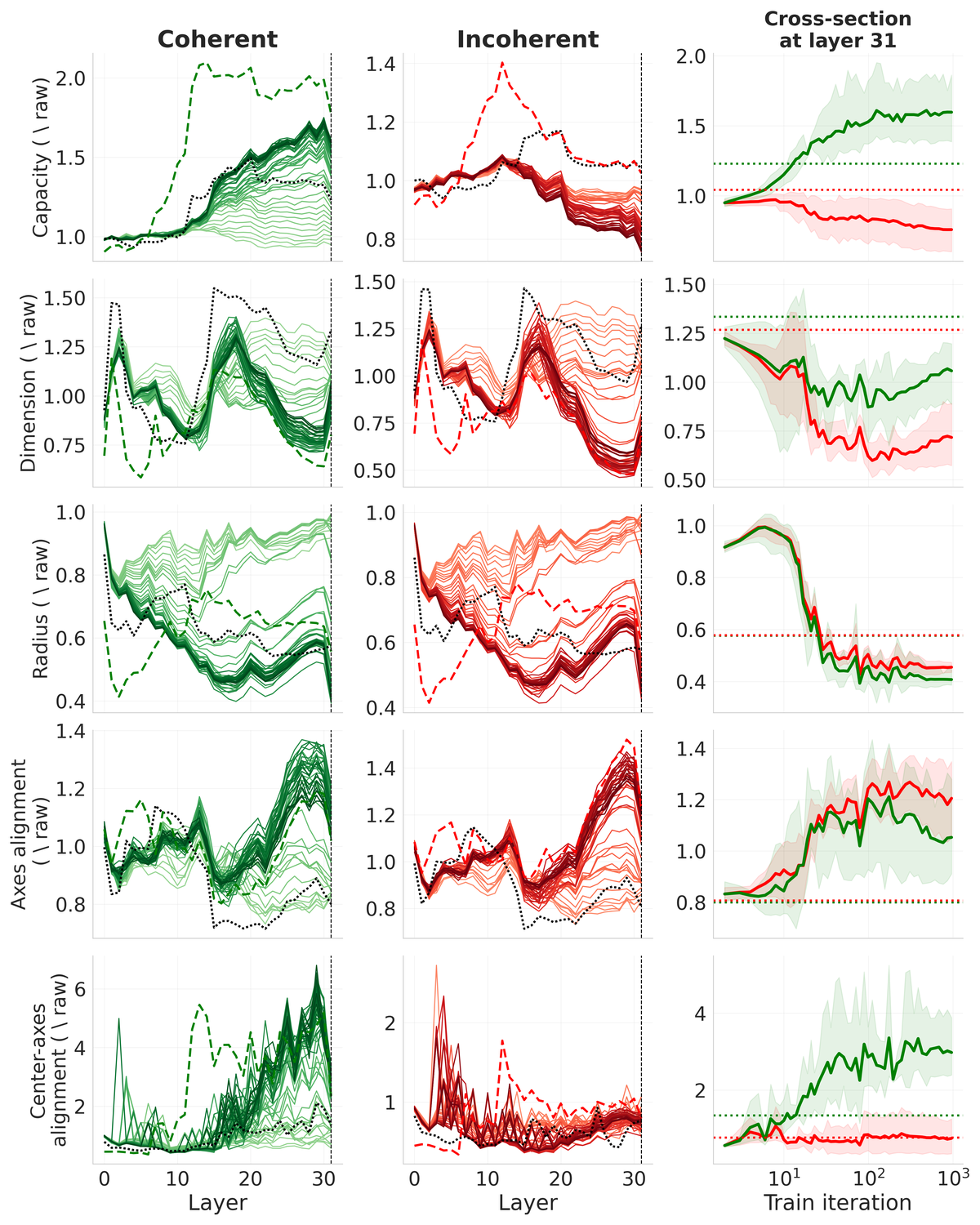}
    \caption{Manifold capacity and geometric measures of \textbf{last token} embeddings during training soft-prompt of length 5 (Llama3.1-8b). Gradient color shows training iterations (darker --- later epochs). Dashed lines --- demonstrations prompt with 40 examples for reference. Dotted --- instruction prompt.}
    \label{fig:prompt_tuning_extended_last_tooken}
\end{figure*}

\end{document}